\documentclass[twoside,11pt]{article}

\usepackage[utf8]{inputenc} 
\usepackage[T1]{fontenc}    
\usepackage{url}            
\usepackage{booktabs}       
\usepackage{amsfonts}       
\usepackage{nicefrac}       
\usepackage{microtype}      
\usepackage{xcolor}         
\usepackage{wrapfig}
\usepackage{multicol}
\usepackage{mathtools}
\usepackage{xcolor}
\usepackage{soul}
\usepackage{todonotes}
\usepackage{wrapfig}
\usepackage[inline]{enumitem}
\usepackage{bbm}
\usepackage{lipsum}

\usepackage{multirow}
\usepackage{adjustbox}
\usepackage{subcaption}
\usepackage{lastpage}
\usepackage[abbrvbib]{jmlr2e}

\usepackage{amsmath, amsfonts, bm}

\def\bigO{\mathcal{O}}
\def\pdff{f}
\def\cdff{\mathcal{F}}

\def\1{\bm{1}}

\def\va{{\bm{a}}}
\def\vb{{\bm{b}}}

\def\ve{{\bm{e}}}

\def\vh{{\bm{h}}}

\def\vp{{\bm{p}}}

\def\vu{{\bm{u}}}
\def\vv{{\bm{v}}}

\def\vx{{\bm{x}}}
\def\vy{{\bm{y}}}
\def\vz{{\bm{z}}}

\def\mA{{\bm{A}}}

\def\mD{{\bm{D}}}
\def\mE{{\bm{E}}}

\def\mH{{\bm{H}}}
\def\mI{{\bm{I}}}

\def\mU{{\bm{U}}}
\def\mV{{\bm{V}}}
\def\mW{{\bm{W}}}
\def\mX{{\bm{X}}}
\def\mY{{\bm{Y}}}
\def\mZ{{\bm{Z}}}

\DeclareMathAlphabet{\mathsfit}{\encodingdefault}{\sfdefault}{m}{sl}
\SetMathAlphabet{\mathsfit}{bold}{\encodingdefault}{\sfdefault}{bx}{n}

\def\gA{{\mathcal{A}}}

\def\gD{{\mathcal{D}}}

\def\gN{{\mathcal{N}}}

\def\gP{{\mathcal{P}}}

\def\gX{{\mathcal{X}}}

\def\sR{{\mathbb{R}}}

\newcommand{\tuple}[1]{{\left\langle #1 \right\rangle}}

\newcommand{\E}{\mathbb{E}}

\newcommand{\sigmoid}{\sigma}

\newcommand{\Var}{\mathrm{Var}}

\newcommand{\Cov}{\mathrm{Cov}}

\DeclareMathOperator*{\argmin}{arg\,min}
\DeclareMathOperator*{\topk}{top-K}
\DeclareMathOperator*{\logsumexp}{\textsc{LogSumExp}}
\DeclareMathOperator*{\argtopk}{arg\,top-K}

\newcommand{\gso}{\mA}
\newcommand{\gsomean}{\boldsymbol{\mu}}
\newcommand{\gsofr}{\gso^\mu}
\newcommand{\ptheta}{{\vp_\theta}}

\newcommand{\frfun}{\mathfrak{F}}
\newcommand{\sensorset}{\mathcal{S}}
\newcommand{\losst}{\mathcal{L}_t\left(\psi,\theta\right)}
\newcommand{\surrlosst}{\widehat{\mathcal{L}}_t\left(\psi,\theta\right)}

\definecolor{ForestGreen}{RGB}{34,139,34}

\jmlrheading{1}{2000}{1-48}{4/00}{10/00}{meila00a}{Marina Meil\u{a} and Michael I. Jordan}

\ShortHeadings{Sparse Graph Learning from Spatiotemporal Time Series}{Cini, Zambon, Alippi}
\firstpageno{1}

\begin{document}

\jmlrheading{24}{2023}{1-\pageref{LastPage}}{10/22; Revised
7/23}{8/23}{22-1154}{Andrea Cini, Daniele Zambon, Cesare Alippi}

\ShortHeadings{Sparse Graph Learning from Spatiotemporal Time Series}{Cini, Zambon and Alippi}

\title{Sparse Graph Learning from Spatiotemporal Time Series}

\author{\name Andrea Cini \email andrea.cini@usi.ch \\
       \addr The Swiss AI Lab IDSIA\\
       Universit\`a della Svizzera italiana\\
       Lugano, CH
       \AND
\name Daniele Zambon \email daniele.zambon@usi.ch \\
       \addr The Swiss AI Lab IDSIA\\
       Universit\`a della Svizzera italiana\\
       Lugano, CH
       \AND
\name Cesare Alippi \email cesare.alippi@usi.ch \\
       \addr The Swiss AI Lab IDSIA\\
       Universit\`a della Svizzera italiana\\
       Lugano, CH\\
       \addr Politecnico di Milano\\
       Milan, IT
}

\editor{Silvia Chiappa}

\maketitle

\begin{abstract}%
Outstanding achievements of graph neural networks for spatiotemporal time series analysis show that relational constraints introduce an effective inductive bias into neural forecasting architectures. Often, however, the relational information characterizing the underlying data-generating process is unavailable and the practitioner is left with the problem of inferring from data which relational graph to use in the subsequent processing stages. We propose novel, principled---yet practical---probabilistic score-based methods that learn the relational dependencies as distributions over graphs while maximizing end-to-end the performance at task. The proposed graph learning framework is based on consolidated variance reduction techniques for Monte Carlo score-based gradient estimation, is theoretically grounded, and, as we show, effective in practice. In this paper, we focus on the time series forecasting problem and show that, by tailoring the gradient estimators to the graph learning problem, we are able to achieve state-of-the-art performance while controlling the sparsity of the learned graph and the computational scalability. We empirically assess the effectiveness of the proposed method on synthetic and real-world benchmarks, showing that the proposed solution can be used as a stand-alone graph identification procedure as well as a graph learning component of an end-to-end forecasting architecture.
\end{abstract}


\section{Introduction}

Traditional statistical and signal processing methods to time series analysis leverage on temporal dependencies to model data generating processes~\citep{harvey1990forecasting}.
Graph signal processing methods extend these approaches to dependencies observed both in time and space, i.e., to the setting where temporal signals are observed over the nodes of a graph~\citep{ortega2018graph, stankovic2020graph2,di2018adaptive, isufi2019forecasting}. The key ingredient here is the use of graph shift operators, constructed from the graph adjacency matrix, that localizes learned filters on the graph structure. The same holds true for graph deep learning methods that have revolutionized the landscape of machine learning for graphs~\citep{bruna2014spectral, bronstein2017geometric, bacciu2020gentle, bronstein2021geometric}. 
However, it is often the case that no prior topological information about the reference graph is available, or that dependencies in the dynamics observed at different locations are not well modeled by the available spatial information~(e.g., the physical proximity of the sensors). Examples can be found in social networks, smart grids, and brain networks, just to name a few relevant application domains. 

The interest in the graph learning problem, in the context of spatiotemporal time series processing, indeed arises from many practical and theoretical concerns. In the first place, learning existing relationships among time series that better explain an observed phenomenon is worth the investigation on its own; as a matter of fact, graph identification is a well-known problem in graph signal processing~\citep{mei2016signal, variddhisai2020methods}. In the deep learning setting, several methods train, end-to-end, a graph learning module with a neural forecasting architecture to maximize performance on the downstream task~\citep{shang2021discrete, wu2020connecting}. A typical deep learning approach consists in exploiting spatial attention mechanisms to discover the reciprocal salience of different spatial locations at each layer~\citep{satorras2022multivariate, rampavsek2022recipe}. Graph learning, in this context, can then be seen as a regularization of Transformer-like models~\citep{vaswani2017attention}; regularization that comes in the form of the relational inductive biases typical of graph processing methods: namely, the sparsity of the pairwise relationships between nodes and the locality of the learned representations. In fact, despite their effectiveness, pure attention-based approaches impair two major benefits of graph-based learning: they (1) do not allow for the sparse computation enabled by the discrete nature of graphs and (2) do not take advantage of the structure, introduced by the graph topology, as an inductive bias for the learning system. Indeed, sparse computation allows graph neural networks~(GNNs;~\citealt{scarselli2008graph}, \citealt{bacciu2020gentle}) with message-passing architectures~\citep{gilmer2017neural} to scale in terms of network depth and the dimension of the graphs that are possible to process. At the same time, sparse graphs constrain learned representations to be localized in node space and mitigate over-fitting spurious correlations in the training data. Graph learning approaches that do attempt to learn relational structures from time series exist, but often rely on continuous relaxations of the binary adjacency matrix and, as a consequence, on dense computations to enable automatic reverse-mode differentiation through any subsequent processing~\citep{shang2021discrete, kipf2018neural}. Conversely, other solutions make the computation sparse~\citep{wu2020connecting, deng2021graph} at the expense of the quality of the gradient estimates as shown by~\citet{zugner2021study}. The challenge is, then, to provide accurate gradients while, at the same time, allowing for sparse computations in the downstream message-passing operations, typical of modern GNNs.

In this paper, we address the graph learning problem and model it from a probabilistic perspective which, besides naturally accounting for uncertainty and the embedding of priors, enables the learning of sparse graphs as realizations of a discrete probability distribution. In particular, given a set of time series, we seek to learn a parametric distribution $\ptheta$ such that graphs sampled from $\ptheta$ maximize the performance on the given downstream task, e.g., multistep-ahead forecasting. As an example, consider a cost function $\delta_t({}\cdot{})$~(e.g., the forecasting accuracy) associated with each time step $t$ and dependant on the inferred graph. The core challenge in learning $\ptheta$ to minimize the expected cost is associated with estimating the gradient
\begin{equation}\label{eq:loss-gradient-intro}
\nabla_\theta \E_{\gso\sim \ptheta}[\delta_t(\gso)]
\end{equation}
of the expected value of the cost function $\delta_t(\gso)$ w.r.t\ the distributional parameters $\theta$, the sampling of a random graph~(adjacency matrix $\gso$) from $\ptheta$ and given batch of input-output data pairs corresponding to observations at time step $t$. 
Previous works proposing probabilistic methods~\citep{shang2021discrete, kipf2018neural} learn $\ptheta$ with \emph{path-wise} gradient estimators~\citep{glasserman1991gradient, kingma2014auto}, i.e., by reparametrizing $\gso\sim\ptheta$ as $\gso=g(\varepsilon,\theta)$, with deterministic function $g$ decoupling parameters $\theta$ from the~(parameter-free) random component $\varepsilon\sim \vp_0$. 
However, these approaches imply approximating discrete distributions with a softmax continuous relaxation~\citep{paulus2020gradient} which makes \emph{all} the downstream computations dense and quadratic in the number of nodes. Differently, here, we adopt the framework of \emph{score-function}~(SF) gradient estimators~\citep{rubinstein1969some, williams1992simple, mohamed2020monte} by relying on the rewriting of Equation~\eqref{eq:loss-gradient-intro} as
\begin{equation}
\nabla_\theta\E_{\gso\sim \ptheta}[\delta_t(\gso)]
= \E_{\gso\sim\ptheta}[\delta_t(\gso)\nabla_\theta \log \ptheta(\gso)]
\end{equation}
which, as we detail in Section~\ref{s:smp}, allows us for preserving the sparsity of the sampled graphs and the scalability of the subsequent processing steps~(e.g., the forward and backward passes of a message-passing network). In particular, our contributions are as follows.
\begin{itemize}

\item We provide an end-to-end methodological framework for probabilistic graph learning in spatiotemporal data, based on SF gradient estimators [Section~\ref{s:lsd}] and design associated Monte Carlo~(MC) estimators for stochastic message-passing architectures [Section~\ref{s:smp}].

\item We introduce two parametrizations of $\ptheta$ as 1) a set of Bernoulli distributions and as 2) the sampling \emph{without} replacement of edges under a sparsity constraint [Section~\ref{s:sampling}]. We show how to sample graphs from both distributions and derive the associated differentiable log-likelihood functions. Both distributions allow us to deal with an adaptive number of neighboring nodes.

\item We propose a novel and effective, yet simple to implement, variance reduction method for the estimators [Section~\ref{s:variance}] based on the Fr\'echet mean graph w.r.t.\ the proposed distributions, for which we provide closed-form solutions~[Propositions~\ref{prop:mu-frechet-bes} and \ref{prop:mu-frechet-sns}].
Our method does not require the estimation of additional parameters and, differently from more general-purpose approaches~(e.g.,  see \citet{mnih2014neural}), is as expensive as taking a sample from the considered distributions and evaluating the corresponding cost function.

\item Finally, we present an approximate surrogate loss function [Section~\ref{s:surrogate}]  derived from a convenient rewriting of the gradient for the considered settings [Proposition~\ref{p:grad}] which provides a considerable improvement in convergence rate.

\end{itemize}

Empirical results demonstrate that the techniques introduced here enable the use of score-based estimators to learn graphs from spatiotemporal time series; furthermore, experiments on time series forecasting benchmarks show that our approach compares favorably w.r.t.\ the state of the art. We strongly believe that our approach constitutes an effective method in the toolbox of the practitioner for designing new, even more effective, classes of novel graph-based time series processing architectures. 
 
The paper is organized as follows. Section~\ref{s:related-works} discusses related works. Section~\ref{s:prelim} introduces relevant background material; Section~\ref{s:problem} provides the formulation of the problem. We present the proposed parametrizations of $\ptheta$ and related gradient estimators in Section~\ref{s:lsd} and the associated variance reduction techniques in Section~\ref{s:variance}. The proposed rewriting of the gradient and approximated objective are derived and discussed in Section~\ref{s:surrogate}. Finally, the empirical evaluation of the proposed method is given in Section~\ref{sec:exp} and conclusions are presented in Section~\ref{s:conclusion}.

\section{Related Works}\label{s:related-works}

Graph neural networks have become increasingly popular in spatiotemporal time series processing~\citep{seo2018structured,li2018diffusion, yu2018spatio, wu2019graph, deng2021graph, cini2022filling, marisca2022learning, wu2021traversenet} and the graph learning problem is well-known within this context.
\citet{wu2019graph} propose Graph WaveNet, an architecture for time series forecasting that learns a  weighted adjacency matrix $\gso=\sigma\left(\mE_1\mE_2^\top\right)$ learned from the factorization with node embedding matrices $\mE_1,\mE_2$. Several other methods follow this direction~\citep{bai2020adaptive, oreshkin2020fcgaga}.
\citet{satorras2022multivariate} showed that hierarchical attention-based architectures are effective to account for dependencies among spatiotemporal time series to obtain accurate predictions in the downstream task.
However, all the aforementioned approaches generally lead to dense graphs and cannot, therefore, exploit the sparsity and locality priors---and computational scalability---typical of graph-based machine learning. 
To address this issue, MTGNN~\citep{wu2020connecting} and GDN~\citep{deng2021graph} sparsify the learned factorized adjacency by selecting, for each node, the $K$ edges associated with the largest weights. Using hard top-k operators, however, results in sparse gradients and has differentiability issues that can undermine the effectiveness of the learning procedure. More recently, \citet{zhang2022graph} proposed a different approach based on the idea of sparsifying the learned graph by thresholding the average of learned attention scores across time steps.

Among probabilistic models, \citet{franceschi2019learning} tackle the graph learning problem for non-temporal data by using a bi-level optimization routine and a straight-through gradient trick~\citep{bengio2013estimating} which, nonetheless, requires dense computations. 
The NRI approach, introduced by \citet{kipf2018neural}, learns a latent variable model predicting the interactions of physical objects by learning edge attributes of a fully connected (dense) graph.
GTS~\citep{shang2021discrete} simplifies the NRI module by considering binary relationships only and integrates graph inference in a spatiotemporal recurrent graph neural network~\citep{li2018diffusion}. 
Both NRI and GTS exploit path-wise gradient estimators based on the categorical \emph{Gumbel trick}~\citep{maddison2017concrete, jang2017categorical} and, as such, rely on continuous relaxations of discrete distributions and suffer from the computational setbacks anticipated in the introduction.
Finally, the graph learning module proposed by \citet{kazi2022differentiable} uses the Gumbel-Top-K trick~\citep{kool2019stochastic} to sample a $K$-nearest neighbors ($K$-NN) graph, where node scores are learned by using a heuristic for increasing the likelihood of sampling edges that contribute to correct classifications. 

Besides applications in graph-based processing, the problem of learning discrete structures has been widely studied in deep learning and general machine learning~\citep{niculae2023discrete}. As alternatives to methods relying on continuous relaxations and path-wise estimators~\citep{jang2017categorical, maddison2017concrete, paulus2020gradient}, several approaches tackled the problem by exploiting score-based estimators and variance reduction techniques, e.g., based on control variates derived from continuous relaxations~\citep{tucker2017rebar, grathwohl2018backpropagation} and data-driven baselines~\citep{mnih2014neural}. In particular, related to our method, \cite{rennie2017self} use a greedy baseline based on the mode of the distribution being learned, while \cite{kool2020estimating} constructs a variance-reduced estimator based on sampling without replacement from the discrete distribution. Beyond score-based and path-wise methods, \cite{correia2020efficient} take a different approach by considering \textit{sparse} distributions where analytically computing the gradient becomes tractable. \cite{niepert2021implicit} introduce a class of (biased) estimators, based on maximum-likelihood estimation, that generalize the straight-through estimator~\citep{bengio2013estimating} to more complex distributions; \cite{minervini2023adaptive} make such estimators adaptive to balance the bias of the estimator and the sparsity of the gradients. We refer to \cite{mohamed2020monte} and \cite{niculae2023discrete} for an in-depth discussion of the topic. None of these method target specifically graph distributions, nor consider sparsity of the downstream computations as a requirement. 

To the best of our knowledge, we are the first to propose a spatiotemporal graph learning module that relies on variance-reduced score-based gradient estimators specifically tailored for graph-based processing, and allowing for sparse computation in both training and inference phases of message-passing neural networks.

\section{Preliminaries}\label{s:prelim}

The section introduces some preliminary concepts and provides the reference models and the notions regarding distributions over graphs needed to support the theoretical and technical derivations presented in the next sections.

\subsection{Spatiotemporal Time Series with Graph Side Information}\label{s:prelim-graphs}

As reference case study, we consider spatiotemporal time series acquired from a sensor network. 
More specifically, consider a set $\sensorset=\{1,2,\dots,N\}$ of $N$ sensors and indicate with $\vx_t^i \in \sR^{d_o}$ the $d_o$-dimensional observation acquired by the $i$-th sensor at discrete time step $t$.
We denote by $\mX_t \in \sR^{N\times d_o}$ the matrix collecting all sensor observations $\{\vx_t^i:i\in\sensorset\}$ at time step $t$. Similarly, whenever available, $\mU_t \in \sR^{N\times d_u}$ indicates the $d_u$-dimensional exogenous variables and with $\mV \in \sR^{N\times d_v}$ static node attributes, e.g., sensor specific features. 
Assume that nodes~(sensors) are available at all time steps and are identified, i.e., a node identifier can be paired to each sensor measurement over time. We also assume node features to be homogeneous across nodes, i.e., to correspond to the same types of sensor readings; an assumption that, however, can easily be relaxed in practice~(e.g., see \citealt{schlichtkrull2018modeling}).

To account for dependencies among measurements at different nodes, observations can be paired with side relational information encoded by an edge set $\mathcal E\subseteq \sensorset\times \sensorset$ or, equivalently, by a (binary) adjacency matrix $\gso\in\{0,1\}^{N\times N}$. Edges of the resulting graph can represent functional dependencies among the different time series that are instrumental for modeling the monitored system and solving the downstream task.
To consider relations that change over time, e.g., as those between users of a social network, we can consider a \emph{dynamic} adjacency matrix $\gso_t$ (or edge set $\mathcal E_t$) representing the variable topology, differently from the \emph{static} case. Finally, $\vy_t \in \sR^{d_y}$ denotes the target vector at every time step, i.e., the task-dependant value to be predicted; targets can also be associated with each node in which case we write $\mY_t \in \sR^{N\times d_y}$.
Often, we are interested in making predictions for a time horizon up to $H$ steps ahead: notation $\mY_{t:t+H}$ denotes the multi-step targets in the interval $[t,t+H)$. Targets define the nature of \emph{downstream task}, which can be either regression or classification, either at the graph or node level. In the following, we consider multi-step node-level tasks as the default setting. 

The above framework is flexible enough to account for several application settings involving sensor measurements; the example below is provided to ease intuition for the reader.
\begin{example} Consider a sensor network monitoring the speed of vehicles at crossroads. In this case, $\mX_{1:T}$ refers to traffic speed measurements sampled at a certain frequency. Exogenous variables $\mU_t$ account for time-of-the-day and day-of-the-week identifiers and, eventually, the current state of traffic lights. The node-attribute matrix $\mV$ reports static features regarding the type of road a sensor is placed in. An adjacency matrix $\mA$ can be obtained by considering each pair of sensors connected if and only if they are connected by a road segment. Targets $\mY_{t}$ provide labels for the task of predicting whether a traffic jam will happen in a fixed number of future time steps or simply one could consider the task of forecasting the next $H$ measurements at each sensor, i.e., $\mY_{t:t+H} = \mX_{t:t+H}$.
\end{example}

\subsection{Spatiotemporal Graph Neural Networks}\label{s:stgn}

The subsection provides an overview of the architectures considered in the sequel. We look at a general class of message-passing operators as well as spatiotemporal graph neural network~(STGNN) architectures.

\subsubsection{Message-Passing Neural Networks} 

We consider the family of message-passing~(MP; \citealt{gilmer2017neural}) operators where representations are updated at each layer $l$ such as

\begin{equation}\label{eq:mp}
    {\vz}^{i, (l)}_t = \rho^{(l)}\Big(\vz^{i, (l-1)}_t, {\textsc{Aggr}}\left\{ \gamma^{(l)}\big( \vz^{j, (l-1)}_t, \vz^{i, (l-1)}_t, \ve_{i,j}\big); j\in\gN(i)\right\}\Big)\\
\end{equation}
where ${\vz}^{i, (l)}_t$ indicates the representation of the $i$-th node at layer $l$; $\gN(i)$ is the set of its neighboring nodes, and $\ve_{i,j}$ are the features associated with the edge connecting the $j$-th to the $i$-th node. Update and message functions, $\rho$ and $\gamma$, respectively, can be implemented by any differentiable function---e.g., a multilayer perceptron---while ${\textsc{Aggr}}\{\cdot\}$ indicates a generic permutation invariant aggregation function. By considering a graph-wise operator, the $l$-th message-passing neural network layer~(MPNN) of the---possibly deep---architecture can be represented in a compact way as
\begin{equation}
    {\mZ}^{(l)}_t = \textsc{MPNN}^{(l)}\left(\mZ_t^{(l-1)}, \mA\right).
\end{equation}

\subsubsection{Spatiotemporal Architectures} 

STGNNs process input spatiotemporal data by considering operators that use the underlying graph to impose inductive biases in the representation learning process. By adopting a terminology similar to the one introduced in~\citep{gao2021equivalence}, we distinguish between time-then-space~(TTS) and time-and-space~(T\&S) STGNNs, depending on whether message-passing is carried out after or in-between a temporal encoding step.

\textit{Time-then-space models.} TTS models are based on an encoder-decoder architecture where the encoder embeds each input time series $\vx^i_{t-W:t}$ associated with a graph node to a vector representation, while the decoder, implemented as a multilayer GNN, propagates information across the spatial dimension. In particular, we consider the family of models s.t.\
\begin{align}
    \mZ^{(0)}_t &= \textsc{TemporalEncoder}\left(\mX_{t-W:t}, \mU_{t-W:t}, \mV\right),\\  
    {\mZ}^{(l)}_t &= \textsc{MPNN}^{(l)}\left(\mZ_t^{(l-1)}, \mA\right),  \quad \forall\ l=1,\dots,L\\
    \widehat \mY_{t:t+H} &= \textsc{Readout}\left( \mZ^{L}_{t} \right),
    \label{eq:tts}
\end{align}
where the notation is consistent with that of Equation~\eqref{eq:mp}. Examples of spatiotemporal graph processing models that fall into the time-then-space category are NRI~\citep{kipf2018neural} and the encoder-decoder architecture introduced by~\citet{satorras2022multivariate}.

\textit{Time-and-space models.} Time-and-space models are a  general class of STGNNs where space and time are processed by operators that process representation along the time and space dimensions. A large subset of this family of models can be seen as performing the following operations
\begin{align}
    \mZ_{t-W:t}^{(0)} &= \left[\mX_{t-W:t} || \mU_{t-W:t} || \mV\right],
\end{align}
then, for every layer $l=1,\dots,L$,
\begin{align}
    \mH^{(l)}_{t-W:t} &= \textsc{TemporalLayer}^{(l)}\left(\mZ^{(l-1)}_{t-W:t}\right),\\ 
    {\mZ}^{(l)}_{k} &= \textsc{MPNN}^{(l)}\left(\mH_{k}^{(l)}, \mA\right),  \qquad \forall\ k=t-W,\dots,t-1\label{eq:stgnn-tts}
\end{align}
finally followed by 
\begin{align}
    \label{eq:pooling-temp}
    \widehat \mY_{t:t+H} &= \textsc{Readout}\left( {\textsc{Aggr}}\left\{ {\mZ}^{(L)}_{t-W}, \dots, {\mZ}^{(L)}_{t-1}\right\} \right),
\end{align}
where $\textsc{TemporalLayer}\left({}\cdot{}\right)$ indicates a generic (parametric) operator processing representations across the different time steps, e.g., a $1$-D convolutional layer.
Note that predictions, here, are obtained in Equation~\eqref{eq:pooling-temp} by pooling representations along the temporal dimension and then using, e.g., a linear readout. Other architectures are possible, e.g., by exploiting recurrent neural networks~\citep{seo2018structured, li2018diffusion}.

\subsection{Mean Adjacency Matrices}\label{s:prelim-frechet}

In this section, we recall some definitions related to probability distributions over graph data. 
The discrete nature of graphs makes a large part of the well-established results from probability and statistics unsuitable for objects that do not adhere to Euclidean geometry. An example is the notion of ``expected'' graph that is of interest to the present paper [Section~\ref{s:variance}] and whose definition needs to be extended. Here, we do so by following \citet{frechet1948elements}.

For a random vector $\vx\in\mathbb{R}^d$ characterized by probability density function $\vp$, expectation 
$\E_{\vx\sim \vp}[\vx] = \int \vx\, \vp(\vx)\,{\rm d}\vx$
is a weighted average over $\vx$; we interchangeably adopt forms $\E_\vp[\vx]$ and $\E[\vx]$. Notably, $\E_{\vx\sim \vp}[\vx]$ can be equivalently written as
\begin{equation}\label{eq:frechet-mean-l2}
\E_{\vx\sim \vp}[\vx] 
= \argmin_{\vx'\in\mathbb R^d} \frfun_{2}(\vx'),
\end{equation}
where $\frfun_{2}({}\cdot{})$ denotes the \emph{Fr\'echet function} 
\begin{equation}\label{eq:frechet-fun-l2}
\frfun_{2}(\vx') \triangleq \E_{\vx\sim \vp}\left[\lVert\vx'-\vx\rVert_2^2\right]
\end{equation}
associated with distribution $\vp$ and the squared Euclidean distance $\lVert\cdot\rVert_2^2$. 
Following Equations~\ref{eq:frechet-mean-l2} and \ref{eq:frechet-fun-l2}, we can derive a generalized definition of mean applicable to non-Euclidean data, like graphs and sparse adjacency matrices. We comment that, following this line, we can extend these results also to the sample mean $1/M \sum_{m=1}^M \vx_m$ of a finite sample $\gD = \{\vx_1,\dots,\vx_M\}$, and define accordingly the Fr\'echet sample mean of a sample of non-Euclidean data.

Consider, then, the space $\gA\subseteq \{0,1\}^{N\times N}$ of adjacency matrices $\gso$ over the node~(sensor) set $\sensorset$, each of which representing a graph topology over $\sensorset$; for instance, for undirected graphs, $\gA$ is the subset of $\{0,1\}^{N\times N}$ of symmetric matrices, whereas for directed $k$-NN graphs 
\begin{equation}\label{eq:support-sns}
\gA = \left\{\gso \in\{0,1\}^{N\times N} : \sum_{j=1}^N \gso_{i,j} = k, \;\forall\,i \right\}.
\end{equation}
By equipping $\gA$ with a metric distance,  we define a Fr\'echet function analogous to that of Equation~\eqref{eq:frechet-fun-l2}, applicable to random adjacency matrices. In this paper, we consider the Hamming distance
\begin{equation}
H(\gso,\gso') \triangleq \sum_{i,j=1}^N I(\gso_{i,j} \ne \gso_{i,j}'),
\end{equation}
where $\gso,\gso'\in \gA$ and $I$ is the indicator function such that $I(a)=1$, if $a$ is true, $0$ otherwise. The Hamming distance counts the number of mismatches between the entries of $\gso$ and $\gso'$, and is then a natural choice to measure the dissimilarity between two graphs. 

We define the Fr\'echet function over space $(\gA,H)$, and the random adjacency matrix $\gso\sim \vp$, for all $\gso'\in\gA$ as
\begin{equation}\label{eq:frechet-fun-H}
\frfun_H(\gso') \triangleq \E_{\gso\sim \vp}\left[H(\gso',\gso)\right].
\end{equation}
According to Equation \ref{eq:frechet-mean-l2}, we then define \emph{Fr\'echet mean adjacency matrix} any matrix
\begin{equation}\label{eq:frechet-mean-H}
\gsofr\in\argmin_{\gso'\in\gA}\frfun_H(\gso'). 
\end{equation}
A matrix $\gsofr$ always exists in $\gA$, as $\gA$ is a finite set, but, in general, is not unique. Conditions for the uniqueness of the Fr\'echet mean in the context of graph-structured data have been studied in the literature, e.g., by \citet{jain2016statistical}. Throughout the paper, we use the term ``Fr\'echet mean'' referring to \emph{any} Fr\'echet mean of a given distribution. 

\section{Problem Formulation}\label{s:problem}

This section provides a probabilistic formulation of the graph learning problem in spatiotemporal time series and defines the operational framework in which we operate. 

\subsection{Graph Learning from Spatiotemporal Time Series} 
Given a window of $W$ past observations ${\gX_{t-W:t} = \tuple{\mX_{t-W:t}, \mU_{t-W:t}, \mV}}$ open on the time series, we consider the problem of predicting $H$ future targets $\mY_{t:t+H}$ associated with the graph nodes. The notation $t:T$ denotes the time steps in interval $[t,T)$; when not differently specified, we consider the multistep-ahead forecasting task $\mY_{t:t+H} = \mX_{t:t+H}$.  

Consider the family of predictive models $F_\psi$ and parametric probability distribution $\ptheta$ over graphs
\begin{equation}\label{eq:generic-model}
    \widehat \mY_{t: t + H} = F_\psi\left(\gX_{t-W:t}\,,\,  \gso_t \right), \quad\quad \gso_t \sim \ptheta\left(\gso \vert \gX_{t-W:t} \right),
\end{equation}
where $\psi$, $\theta$ are the model parameters. The joint graph and model learning problem consists in jointly learning parameters $\psi$, $\theta$ by solving the optimization problem
\begin{equation}\label{eq:generic-losses}
\hat \psi,\hat \theta = \argmin_{\psi,\theta}\frac{1}{T}\sum_{t=1}^{T}\losst,
 \qquad\quad 
\losst  \triangleq \E_{\gso\sim\ptheta}\left[\delta_t(\gso_t; \psi)\right], 
\end{equation}
where $\losst$ is the  optimization objective at time step $t$ expressed as the expectation, over the graph distribution $\ptheta$, of a cost---loss---function $\delta_t(\gso_t; \psi)$, typically a $p$-norm
\begin{equation}
    \delta_t(\gso_t; \psi) = \lVert \mY_{t:t+H} - F_\psi\left(\gX_{t-W:t},  \gso_t \right)\rVert_p^p    
\end{equation}
with, e.g., $p=1$ or $2$. Note that in Equation~\eqref{eq:generic-model} the distribution of $\gso_t$ at time step $t$ is conditioned on the most recent observations $\gX_{t-W:t}$, hence modeling a scenario associated with a dynamic graph distribution [Section~\ref{s:prelim-graphs}]. A static graph scenario follows by simply removing the conditioning on $\gX_{t-W:t}$. 
We consider a generic family of predictive models $F_\psi$ implemented by STGNNs based on the message-passing framework and following either the TTS or the T\&S paradigm to process information along space and time. Other architectures can be considered. Notably, $F_\psi$ can be suitably designed in order to exchange messages w.r.t.\ a different graph $\gso^{(l)}$ at each MP layer. Section~\ref{s:surrogate} provides a thorough discussion of this setup. 

In this setting, the model family and the downstream task impact on the type of relationships being learned. For example, linear and nonlinear models will yield different results that depend also on the number of layers and the choice of MP operators, e.g., standard graph convolutions against anisotropic message-passing layers such as those used in graph attention networks~\citep{velivckovic2018graph}. Ultimately, the learned graph distribution is the one that best explains the observed data given the architecture of the predictive model and the family of graph distributions. Different parametrizations of $\ptheta$ allow the practitioner for embedding different inductive biases~(such as sparsity) as structural priors into the processing.

\subsection{Core Challenge}
Minimizing the sum of expectations $\losst$, $t=1,\dots,T$, is challenging, as it involves estimating the gradients $\nabla_\theta \losst$ w.r.t.\ the parameters of the discrete distribution $\ptheta$ over (binary) adjacency matrices. Sampling matrices (graphs) $\gso\sim \ptheta$ throughout the learning process results in a stochastic computational graph (CG) and, while automatic differentiation of CGs is a core component of modern deep learning libraries~\citep{paske2019pytorch, abadi2015tensorflow}, dealing with stochastic nodes introduces additional challenges as the gradients have to be estimated w.r.t.\ expectations over the sampling of the associated random variables~\citep{schulman2015gradient, weber2019credit, mohamed2020monte}. Tools for automatic differentiation of stochastic CGs are being developed~\citep{foerster2018dice, bingham2019pyro, krieken2021storchastic, dillon2017tensorflow}; however, general approaches can be ineffective and prone to failure, especially in the case of discrete distributions~(see also~\citealt{mohamed2020monte}). 

In our setup, having a stochastic message-passing graph~(MPG) emerges as problematic: the MP paradigm constrains the flow of spatial information, making the CG dependent on the MPG.  Moreover, a stochastic input MPG introduces $N^2$ stochastic nodes in the resulting CG~(i.e., one for each potential edge in MPG), leading to a large number of paths data can flow through. For instance, by considering an $L$-layered architecture, the number of stochastic nodes can increase up to $\bigO(LN^2)$, making the design of reliable, low-variance---i.e., effective---MC gradient estimators inherently challenging. Furthermore, as already mentioned, computing gradients associated with each stochastic edge introduce additional challenges w.r.t.\ time and space complexity; further discussion and actionable directions are given in the next section.

\section{Score-based Sparse Graph Learning from Spatiotemporal Time Series}\label{s:lsd}

\begin{figure}
    \centering
    \includegraphics[scale=0.8]{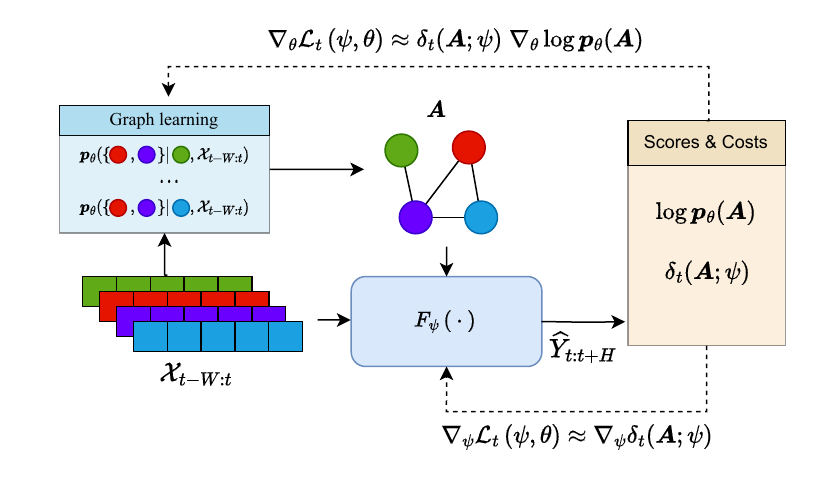}
    \caption{Overview of the learning architecture. The graph learning module samples a graph used to propagate information along the spatial dimension in $F_\psi$; predictions and samples are used to compute costs and log-likelihoods. Gradient estimates are propagated back to the respective modules.}
    \label{fig:framework}
\end{figure}

In this section, we present our approach to probabilistic graph learning. After introducing score-based gradient estimators [Section~\ref{s:smp}], we propose two graph distribution models~[Section~\ref{s:sampling}] and comment on their practical implementations [Section~\ref{s:ptheta}]. The problem of controlling the variance of the estimators is discussed together with novel and principled variance reduction techniques tailored to graph-based architectures [Section~\ref{s:variance}]. Finally, we provide a convenient rewriting of the gradient for $L$-layered MP architectures leading to a novel surrogate loss~[Section~\ref{s:surrogate}]. Figure~\ref{fig:framework} provides a schematic overview of the framework. In particular, the block on the left shows the graph learning module, where $\gso$ is sampled from $\ptheta$; as the figure suggests, depending on the parametrization of~$\ptheta$, some components of $\gso$ can be sampled independently. The bottom of the figure, instead, shows the predictive model $F_\psi$ that, given the sampled graph and the input window, outputs the predictions used to estimate $\losst$, whose gradient provides the learning signals.

\subsection{Estimating Gradients for Stochastic Message-Passing Networks}\label{s:smp}

SF estimators are based on the identity 
\begin{align}
\nabla_\theta\E_{\ptheta}[f(x)] &= \nabla_\theta \int f(x) \ptheta(x) \,d x = \int f(x) \nabla_\theta \ptheta(x) \,d x\label{eq:interchange}\\ 
&=\int f(x) \ptheta(x) \nabla_\theta\log\ptheta(x) \,d x=\E_{\ptheta}[f(x)\nabla_\theta \log \ptheta(x)],\label{eq:score-function-identity}
\end{align}
which holds---under mild assumptions\footnote{The identity is valid as long as $\ptheta$ and $f$ allow for the interchange of differentiation and integration in Equation~\eqref{eq:interchange}; see~\citet{l1995note, mohamed2020monte}.}---for generic cost functions $f$ and distributions $\ptheta$. The rewriting of $\nabla_\theta\E_{\ptheta}[f(x)]$ in terms of the gradient of the \emph{score function} $\log\ptheta({}\cdot{})$ allows for estimating the gradients easily by MC sampling and backpropagating them through the computation of the score function. SF estimators are black-box optimization methods, i.e.,  they only require to \emph{evaluate} pointwise the \emph{cost} $f(x)$ which does not necessary need to be differentiable w.r.t.\ parameters $\theta$. In our setup, by assuming disjoint $\psi$ and $\theta$, Equation~\eqref{eq:score-function-identity} becomes
\begin{equation}\label{eq:score-function-identity-graph}
\nabla_\theta\losst 
= \nabla_\theta\E_{\ptheta}\left[\delta_t(\gso;\psi)\right] = \E_{\ptheta}\left[\delta_t(\gso;\psi)\nabla_\theta \log \ptheta(\gso)\right],
\end{equation} 
allowing for computing gradients w.r.t.\ the graph generative process without requiring a full evaluation of all the stochastic nodes in the CG. 

\textit{Sparse computation.} Path-wise gradient estimators tackle the problem of estimating the gradient $\nabla_\theta\E_{\ptheta}\left[\delta_t(\gso;\psi)\right]$ by exploiting continuous relaxations of the discrete $\ptheta$, thus estimating the gradient by differentiating through all nodes of the stochastic CG.
Defined $E$ to be the number of edges in a realization of $\ptheta$, the cost of learning a graph with a path-wise estimator is that of making any subsequent MP operation scale with $\bigO(LN^2)$, instead of the $\bigO(LE)$ complexity that would have been possible with a sparse computational graph. The outcome is even more dramatic if we consider T\&S models where MP is used for propagating information at each time step, thus making the computational and memory complexity $\bigO(LTN^2)$, which would be unsustainable for any practical application at scale. Conversely, the proposed score-based methods allow for the implementation of MP operators with efficient scatter-gather operations that exploit the sparsity of $\gso$, thus resulting in an $\bigO(LE)$ complexity.

\subsection{Graph Distributions, Graphs Sampling, and Graphs Likelihood}\label{s:sampling}

The distribution $\ptheta$ should be chosen to (i) efficiently sample graphs and evaluate their likelihood and (ii) backpropagate the errors through the computation of the score~[Equation~\eqref{eq:score-function-identity-graph}] to parameters $\theta$.
In the following, we  consider graph distributions s.t.\ each stochastic edge $j\to i$ is associated with a weight $\Phi_{i,j}$. 
The considered distributional parameters $\Phi \in \sR^{N\times N}$ can then be learned as a function of the learnable parameters $\theta$.
In the case of static graphs, we can directly consider $\Phi=\theta$; however, to account for the dynamic case, more complex parametrizations are possible, e.g., by exploiting amortized inference to condition distribution $\ptheta$ on the observed values. Further discussion is deferred to the end of the section. 

\subsubsection{Binary Edge Sampler}
A straightforward approach considers a Bernoulli random variable of parameter $\sigmoid(\Phi_{i,j})$ associated with each potential edge $j\to i$. We refer to this graph learning module as \emph{binary edge sampler}~(BES).

\textit{Sampling.} 
For all pairs of sensors $i,j\in\sensorset$, the corresponding entries $\gso_{i,j}$ of $\gso$ can be sampled independently from the associated distribution since $\gso_{i,j} \sim \text{Bernoulli}(\sigmoid(\Phi_{i,j}))$. Here, the sampling from $\ptheta$ can be done efficiently and is highly parallelizable.

\textit{Log-likelihood evaluation.} Computing the log-likelihood of a sample is cheap and differentiable as it corresponds to evaluating the binary cross-entropy between the sampled entries of $\gso$ and the corresponding parameters $\sigma(\Phi)$ of the Bernoulli distributions, i.e,
\begin{equation}
\log \ptheta(\gso) = \sum_{i,j}^N \gso_{i,j} \log(\sigma(\Phi_{i,j})) + (1 - \gso_{i,j}) \log(1 - \sigma(\Phi_{i,j})).   
\end{equation}

Sparsity priors can then be imposed by regularizing $\Phi$, e.g., by adding a Kullback-Leibler regularization term to the loss~\citep{shang2021discrete, kipf2018neural}. 
Graph generators like BES are a common choice in the literature~\citep{franceschi2019learning, shang2021discrete} as the independence assumption makes the mathematics amenable and avoids the often combinatorial complexity of dealing with more structured distributions. In the experimental sections, we demonstrate that even simple parametrizations like BES can be effective with the proposed score-based learning.

\subsubsection{Subset Neighborhood Sampler}\label{s:sns}
Encoding structural priors about the sparseness of the graphs directly into $\ptheta$ is often desirable and might allow---depending on the problem---to remarkably reduce sample complexity. In this section, we use the score matrix $\Phi\in\mathbb{R}^{N\times N}$ to parametrize a stochastic top-k sampler that we dub \emph{subset neighborhood sampler}~(SNS). 
For each $n$-th node, we sample a subset $S_K\subset \sensorset = \{1,\dots,N\}$ of $K$ neighboring nodes by sampling \emph{without replacement}  from a categorical distribution parametrized by \emph{normalized log-probabilities} $\Phi_{n,:}$. 
The probability of sampling neighborhood $S_K$ for each node $n$ is given by
\begin{equation}\label{e:unord}
    \ptheta(S_K|n)=\sum_{\vec S_K \in \mathcal{P}(S_K)}\ptheta(\vec S_K|n) 
    =\sum_{\vec S_K \in \mathcal{P}(S_K)}\prod_{j\in\vec S_K} \frac{\exp(\Phi_{n,j})}{1-\sum_{k<j}\exp(\Phi_{n,k})},
\end{equation}
where $\vec S_K$ denotes an ordered sample without replacement and $\gP{(S_K)}$ is the set of all the permutations of $S_K$. 

\textit{Sampling.}
Sampling can be done efficiently by exploiting the \emph{Gumbel-top-k trick}~\citep{kool2019stochastic}. Accordingly, we consider the parameter vector $\phi=\Phi_{n,:}$ and denote with $[G_{\phi_1},\dots,G_{\phi_N}]$ the associated random vector of independent Gumbel random variables $G_{\phi_j}\sim \text{Gumbel}({\phi_j})$; given a realization thereof $[g_1,\dots,g_N]$, it is possible to show that $S_K=\argtopk\{g_i:i\in \sensorset\}$ follows the desired distribution~\citep{kool2019stochastic}. 

\textit{Log-likelihood evaluation.}
Evaluating the score function is more challenging; in fact, Equation~\eqref{e:unord} shows that directly computing $\ptheta(S_K|n)$ requires marginalizing over all the possible $K!$ orderings of $S_K$. While exploiting the Gumbel-max trick can bring down computation to $\bigO(2^K)$~\citep{huijben2022review, kool2020estimating}, exact computation remains untractable for any practical application. Luckily, $\ptheta(S_K|n)$ can be approximated efficiently using numerical integration. Following the notation of~\citet{kool2019stochastic, kool2020estimating}, for a subset $B\in\sensorset$ we define 
\begin{equation}\label{e:logsumexp}
\logsumexp_{i\in B} (\phi_i) \triangleq \log \left(\sum_{i\in B}\exp\phi_i \right),
\end{equation}
we use the notation $\phi_B = \logsumexp_{i\in B} \phi$, and indicate with $\pdff_u$ and $\cdff_u$ the p.d.f.\ and c.d.f., respectively, of a Gumbel random variable $\text{Gumbel}(u)$ with location parameter $u$. Recall that $\cdff_u(z) = \exp(-\exp(-z + u))$ and the following property of Gumbel random variables:
\begin{equation}\label{e:gmax}
    G_{\phi_B} \triangleq \max_{i\in B} G_{\phi_i} \sim \text{Gumbel}(\phi_B).
\end{equation}
With a derivation analogous to that of~\citet{kool2020estimating}, Equation~\eqref{e:unord} can be conveniently rewritten by exploiting the property shown in Equation~\eqref{e:gmax} as:
\begin{align}
    \ptheta(S_K|n) &= \mathbb{P}\left(\min_{i\in S_K} G_{\phi_i} > \max_{i\in {\sensorset\setminus S_k}} G_{\phi_i}\right)\\
    &= \mathbb{P}\left(G_{\phi_i} > G_{\phi_{\sensorset \setminus S_k}}, \forall i \in S_K\right)\\
    &= \int_{-\infty}^{\infty} \prod_{i\in S_K}\left(1 - \cdff_{\phi_i}\left(g\right)\right)\pdff_{\phi_{\sensorset \setminus S_k}}(g) \,d g
\end{align}
With an appropriate change~(details in Appendix~\ref{a:trapezoid}), the integral can be rewritten as 
\begin{equation}\label{eq:sns-score-rewriting}
    \ptheta(S_K|n) = \exp\left( \phi_{\sensorset \setminus S_K} + c\right)\int_0^1 u^{\exp\left(\phi_{\sensorset \setminus S_K} + c\right) - 1}\prod_{i\in S_k}\left(1 - u^{\exp(\phi_i + c)}\right)\,du,
\end{equation}
where $c$ is a conditioning constant. We then approximate the integral in Equation~\eqref{eq:sns-score-rewriting}  by using the trapezoidal rule as 
\begin{multline}\label{eq:sns-apporx-score}
    \log{\ptheta(S_K|n)} \approx
    \log(\Delta u) + \phi_{\sensorset \setminus S_K} + c \\ +
    \logsumexp_{m=1,\dots,M-1}\left(\Big(\exp\left(\phi_{\sensorset \setminus S_K} + c\right) - 1\Big)\log(u_m) + \sum_{i\in S_K}\log\left(1-u_m^{\exp(\phi_i + c)}\right)\right),
\end{multline}
with $M$ trapezoids and equally spaced intervals of length $\Delta u$; the integrands are computed in log-space---with a computational complexity of $\bigO(MK)$---for numeric stability. The expression in Equation~\eqref{eq:sns-apporx-score} provides, then, a differentiable numeric approximation of the SNS log-likelihood which can be used for backpropagation. 

As previously discussed, the proposed SNS method allows for embedding structural priors on the sparsity of the latent graph directly into the generative model. Fixing the number $K$ of neighbors might, however, introduce an irreducible approximation error when learning graphs with nodes characterized by a variable number of neighbors. We solve this problem by adding dummy nodes.

\textit{Adaptive number of neighbors.} 
Given $K$, we add up to $K-1$ dummy nodes to set $\sensorset$~(i.e. the set of candidate neighbors) and expand matrix $\Phi$ accordingly. At this point, a neighborhood of exactly $K$ nodes can be sampled and the log-likelihood evaluated according to the procedure described above; however, dummy nodes are discarded to obtain the $N\times N$ adjacency matrix $\gso$. By doing so, hyperparameter $K$ can also be used to cap the maximum number of edges and set a minimum sparsity threshold. The resulting computational complexity in the subsequent MP layers is at most $\bigO(NK)$.

\subsection[Parametrizing]{Learning the Parameters of the Graph Distribution $\ptheta$}\label{s:ptheta}

As previously mentioned, for both BES and SNS, we can parametrize $\ptheta$ by associating a score $\Phi_{i,j}$ to each edge $j\to i$; i.e., by setting $\Phi = \theta$.  
Similarly, one could reduce the number of parameters to estimate from $N^2$ to $2dN$, with $d\ll N$, by using amortized inference and learning some factorization of $\Phi$, e.g., $\Phi = \theta_s \theta_t^\top$ where $\theta_s, \theta_t \in \sR^{N\times d}$~(e.g., see~\citealt{kipf2016variational, kipf2018neural}). Modeling dynamic graphs instead requires accounting for observations $\gX_{t-W:t}$ at each considered time step $t$. For example, one can consider models s.t.\
 \begin{equation}\label{e:phi}
     \vh^i_t = \textsc{Encoder}\left(\vx^i_{t-W: t}, \vu^i_{t-W: t}, \vv^i\right), \qquad \phi_{i,j} = \va^\top\sigma\left(\mW [\vh^i_t \vert\vert \vh^j_t] + \vb\right),
 \end{equation}
 where $\textsc{Encoder}(\cdot)$ indicates a generic encoding function for in the input window (e.g., an MLP or an RNN), $\sigma$ a nonlinear activation function, $\mW\in \sR^{d\times 2d_h}$ is a learnable weight matrix, $\vb\in d$ a learnable bias and $\va \in \sR^{d}$ the learnable parameters of the output linear transformation.

\section{Reducing the Variance of the Estimator}\label{s:variance}

MC estimation is the most commonly used technique to approximate the gradient in Equation~\eqref{eq:score-function-identity-graph}. Although MC estimators are unbiased, the quality of the estimate can be dramatically impacted by its variance:
as such, variance reduction is a critical step in the use of score-based estimators. As for any MC estimator, a direct method to reduce the variance consists in increasing the number $M$ of independent samples used to compute the estimator, which results in reducing the variance by a factor $1/M$ w.r.t.\ the one-sample estimator. In our setting, sampling $M$ adjacency matrices results in $M$ evaluations of the cost and the associated score and, in turn, to an often non-negligible computational overhead. In this section, we provide more sample-efficient alternatives, based on the control variates method. Our approach grants a significant variance reduction while requiring only one extra evaluation of the cost function. That being said, our approach to variance reduction is orthogonal to increasing the sample size, which remains viable to further improve the quality of the gradient estimator.

\subsection{Control Variates and Baselines} 

The control variates method provides a variance reduction method for MC estimator of $\E_\ptheta[g(x)]$. It consists in introducing an auxiliary quantity $h(x)$ for which we know how to efficiently compute the expectation under the sampling distribution $\ptheta$~\citep{mohamed2020monte}. Then, a function ${\tilde g=g-\beta(h - \E[h])}$ is defined, for some constant $\beta$, such that $\tilde g$ has the same expected value of $g$, i.e., $\E[\tilde g(x)]=\E[g(x)]$, but lower variance~($\Var[\tilde g(x)]<\Var[g(x)]$). Quantity $h$ is called \emph{control variate}, while $\beta$ is often referred to as \emph{baseline}.
In score-based methods, a computationally cheap choice is to use the score function itself as control variate, i.e., referring to our case where $g(\gso) \triangleq \delta_t(\gso;\psi)\nabla_\theta \log \ptheta(\gso)$~(Equation~\eqref{eq:score-function-identity-graph}), we set $h(\gso) \triangleq \nabla_\theta\log\ptheta(\gso)$, for which $\E_\ptheta[h(\gso)]=0$, and obtain
\begin{equation}\label{eq:score-function-identity-baseline}
\nabla_\theta\losst
= \E_{\ptheta}\left[\left(\delta_t(\gso;\psi)-\beta\right)\nabla_\theta \log \ptheta(\gso)\right].
\end{equation}
This narrows the problem to finding appropriate values for baseline $\beta$.

Since for any $f_1,f_2$, $\Var[f_1+f_2]=\Var[f_1] + \Var[f_2] + 2\Cov[f_1,f_2]$, the optimal baseline $\beta_*$ in Equation~\eqref{eq:score-function-identity-baseline} is given by
\begin{equation}\label{eq:beta-star}
\beta_*
\triangleq\frac{\Cov[\delta_t(\gso;\psi)\nabla_\theta\log\ptheta(\gso),\nabla_\theta\log\ptheta(\gso)]}{\Var_\ptheta[\nabla_\theta\log\ptheta(\gso)]}
=\frac{\E_\ptheta[\delta_t(\gso;\psi)(\nabla_\theta\log\ptheta(\gso))^{ 2}]}{\E_\ptheta[(\nabla_\theta\log\ptheta(\gso))^2]}. 
\end{equation}
Unfortunately, finding the optimal $\beta_*$ can be as hard as estimating the desired gradient in Equation~\eqref{eq:score-function-identity-graph}; moreover, note also that $\beta_*=\beta_*(\gX_t)$, as $\delta_t$ depends on the observations $\gX_t$. 

Therefore, we opt for the approximation  
\begin{align}
    \E_\ptheta[\delta_t(\gso;\psi)(\nabla_\theta\log\ptheta(\gso))^{ 2}] 
    &\approx 
    \E_\ptheta[\delta_t(\gso;\psi)]\E_\ptheta[(\nabla_\theta\log\ptheta(\gso))^{ 2}],
\end{align}
and obtain $\beta_*\approx\E_\ptheta[\delta_t(\gso;\psi)]$. Note that a similar choice of baseline is very popular, for instance, in reinforcement learning applications~(e.g., see advantage actor-critic estimators,~\citealt{sutton1999policy, mnih2016asynchronous}). However, since approximating $\E_\ptheta[\delta_t(\gso;\psi)]$ would require the introduction of an additional estimator, we rely on a different approximation by moving the expectation inside the cost function and obtaining $\beta_*\approx\delta_t(\gsomean;\psi)$, where $\gsomean=\E_\ptheta[\gso]$.

We recall that, in general, $\gsomean$ is dense and its components are real numbers, therefore computing $\delta_t(\gsomean;\psi)$ would require evaluating the output of the model w.r.t.\ a dense adjacency matrix, potentially outside the well-behaved region of the input space, and to compute messages w.r.t.\ each node pair, thus negating any computational complexity benefit. Accordingly, we substitute $\gsomean$ with the Fr\'echet mean adjacency matrix $\gsofr$, relying on the generalized notion of mean for binary adjacency matrices introduced in Section~\ref{s:prelim-frechet}. We then choose as $\hat\beta$ such that 
\begin{equation}\label{eq:beta-hat}
\hat\beta \triangleq \delta_t(\gsofr;\psi).
\end{equation}
The computational cost of evaluating $\hat\beta$ corresponds then to that of a single evaluation of the cost function $\delta_t$ w.r.t.\ the binary and eventually sparse adjacency matrix $\gsofr$.

Finally, we point out that, even though $\hat \beta$ may differ from $\beta_*$, the variance is reduced as long as $0<\hat\beta< 2\beta_*$. We indicate the modified cost, i.e., the cost minus the baseline as $\tilde{\delta}_t(\gso;\psi)=\delta_t(\gso;\psi) - \delta_t(\gsofr;\psi)$; the modified cost is computed after each forward pass and used to update the parameters of $\ptheta$.
In next Sections~\ref{sec:frechet-mean-bes} and \ref{sec:frechet-mean-sns} we derive analytic solutions for finding $\gsofr$ for BES and SNS, respectively.

\subsection{Baseline for BES}\label{sec:frechet-mean-bes}

We start by recalling the notation from previous sections. Denote expectation $\E_\ptheta[\gso]$ with respect to BES as $\gsomean\in [0,1]^{N\times N} \subset \sR^{N \times N}$ and with $\gsofr$ the binary Fr\'echet mean adjacency matrix with respect to the support $\gA=\{0,1\}^{N\times N}$ of the distribution $\ptheta$ associated with BES. The main result of the section is the following proposition which allows us to provide a baseline as
\begin{equation}
    \hat \beta_\text{BES} \triangleq \delta_t\left(\lfloor \sigma(\Phi)\rceil;\psi\right),
\end{equation}
where $\lfloor\Phi\rceil$ indicates the element-wise rounding of the components of the real matrix $\Phi$ to the closest integer~(either $0$ or $1$).

\begin{proposition}\label{prop:mu-frechet-bes}
Consider BES with associated distribution $\ptheta$ and support $\gA$. Then,
\begin{itemize}
    \item the expected matrix $\E_\ptheta[\gso]$ is $\gsomean=\sigmoid(\Phi)$, with $\sigmoid$ applied element-wise;
    \item the Fr\'echet mean adjacency matrix $\gsofr=\lfloor \gsomean\rceil = I(\Phi > 0)$. 
\end{itemize}
\end{proposition}

\medskip
\begin{proof}
As each component of $\gso\sim \ptheta$ is independent from the others,  $\gsomean_{i,j}$ can be considered element-wise as $\E_\ptheta[\gso_{i,j}]=\sigma(\Phi_{i,j})$, for all $i,j=1,\dots,N$.
Similarly, each component of $\gsofr$ can be computed independently as well, by relying on Lemma~\ref{lemma:frechet-fun}. 
\begin{lemma}\label{lemma:frechet-fun}
The minimum of the Fr\'echet function $\frfun_H$ can be expressed as  
\begin{equation}\label{eq:frechet-minimum}
\min_{\gso \in\gA}\frfun_H(\gso) = \min_{\gso \in\gA}\sum_{i,j=1}^N \left(\gsomean_{i,j}-\gso_{i,j} \right)^2.
\end{equation}
\end{lemma}
To conclude the proof of Preposition~\ref{prop:mu-frechet-bes}, we observe that the minimum of Equation~\eqref{eq:frechet-minimum} is attained at $\gsofr=\lfloor\gsomean\rceil$, that is $\gsofr_{i,j} = 1$ for all $\gsomean_{i,j}>1/2$~(or $\Phi > 0$), and $0$ elsewhere. The proof of the Lemma~\ref{lemma:frechet-fun} is deferred to Appendix~\ref{a:proofs}.
\end{proof}

\subsection{Baseline for SNS}\label{sec:frechet-mean-sns}

Similarly to what has been done for BES in Proposition~\ref{prop:mu-frechet-bes}, we provide analogous results for SNS, with the added technical complexity that, in this case, edges $j\to i$ and $j'\to i$ are not independent. Nevertheless, the result remains intuitive:
\begin{equation}\label{eq:SNS-baseline}
    \hat \beta_\text{SNS} \triangleq \delta_t\left(\gsofr;\psi\right), \quad\text{ with } \gsofr_{i,j} = I\left(\Phi_{i,j} \in \small{\topk} \{\Phi_{i,:}\}\right),\ \forall\;i,j\in\sensorset.
\end{equation}
The proof that $\gsofr$ is indeed the Fr\'echet mean for SNS follows Preposition~\ref{prop:mu-frechet-sns}.
Recall that, for SNS, the support of $\ptheta$ is that of directed $K$-NN graphs in Equation~\eqref{eq:directed-knn-set}, where the neighborhood of each node is sampled independently. Equation~\eqref{eq:SNS-baseline} is derived by considering a neighborhood of fixed size $K$; however, the analysis remains valid for the adaptive case discussed in Section \ref{s:sns}.

In the SNS case, each entry $\gsomean_{n,i}$ of $\gsomean$ is  
\begin{equation}
    \gsomean_{n,i} = \ptheta(i \in S_K | n)=\sum_{S_K':\, i\, \in\, S_K'} \ptheta(S_K' | n),    
\end{equation}
where the sum is taken over all subsets $S_K'$ of $\sensorset$ of $K$ elements containing node $i$.
Even if marginalizing over all possible sampled subsets of $S_K'$ has combinatorial complexity, we show that $\gsofr$ can be derived without directly computing $\gsomean$ as stated in Proposition~\ref{prop:mu-frechet-sns}.
\begin{proposition}\label{prop:mu-frechet-sns}
Consider an SNS distribution with support 
\begin{equation}\label{eq:directed-knn-set}
\gA = \left\{\gso \in\{0,1\}^{N\times N} : \small{\sum}_{j=1}^N \gso_{i,j} = K, \;\forall\,i \right\}.
\end{equation}
Then, the Frechét mean $\gsofr$ is given by
\begin{equation}
    \gsofr_{i,j} = I\left(\Phi_{i,j} \in \small{\topk} \{\Phi_{i,:}\}\right),\ \forall\;i,j\in\sensorset.
\end{equation}
\end{proposition}
\begin{proof}
Computing $\gsofr$ corresponds to solving the optimization problem
\begin{align}\label{eq:SNS-fr-obj}
    \min_{\gso \in \gA} \frfun_H\left(\gso\right) = \min_{\gso \in \gA} \E_{\gso^\prime \sim \ptheta} \left[H(\gso, \gso^\prime)\right].
\end{align}
Start by rewriting the Fr\'echet function as
\begin{align}
     \frfun_H(\gso) &= \E_{\gso^\prime \sim \ptheta} \left[H(\gso, \gso^\prime)\right]\\ 
     &= \E_{\gso^\prime \sim \ptheta} \left[\sum_{n,i=1}^N\gso_{n,i} - 2\gso_{n,i}\gso_{n,i}^{\prime}  + \gso_{n,i}^{\prime}\right]\\ 
     &= \sum_{n,i=1}^N\gso_{n,i} \underbrace{(1 - 2\gsomean_{n,i}')}_{w_{n,i}}  + \underbrace{\sum_{n,j=1}^N \gsomean_{n,i}}_{c}.\label{eq:frfun-sns-lin}
\end{align}
where $\gsomean_{n,i}=\ptheta\left(i \in S_K | n\right) = \ptheta\left(\gso_{n,i}=1\right)$ and $c$ is a constant. The proof follows from Lemma~\ref{lemma:pi-geq-pj}.
\begin{lemma}\label{lemma:pi-geq-pj}
    Let $\ptheta$ be an SNS distribution with associated log-probabilities $\Phi$.
    Then $\forall n,i,j\in \sensorset$
    \begin{equation}
        \ptheta\left(\gso_{n,i}=1\right) \geq \ptheta\left(\gso_{n,j}=1\right) \iff \Phi_{n,i} \geq \Phi_{n,j}.
    \end{equation}
\end{lemma}
The proof of Lemma~\ref{lemma:pi-geq-pj} is provided in Appendix~\ref{a:proofs}. Following Equation~\eqref{eq:frfun-sns-lin}, the optimization problem in Equation~\eqref{eq:SNS-fr-obj} becomes the linear program
\begin{equation}\label{eq:sns-linear-program}
    \begin{aligned}
    \text{minimize} &\sum_{i=1}^N \sum_{j=1}^N w_{i,j} \gso_{i,j}\\
    \text{s.t.\ } &\sum_{j=1}^N \gso_{i,j} = K;\\
    & \gso_{i,j} \in \{0,1\} \qquad \forall i=1,\dots,N,\\
    \end{aligned}
\end{equation}
where $w_{i,j} = 1 - 2\ptheta\left(\gso_{i,j}=1\right)$.
Since Lemma~\ref{lemma:pi-geq-pj} grants that, for each $i$, the $K$-smallest $w_{i,j}$ weights correspond row-wise to the top-$K$ scores $\Phi_{i,j}$, the solution $\gsofr$ to the linear program is given by $\gsofr_{i,j} = I\left(\Phi_{i,j} \in \small{\topk} \{\Phi_{i,:}\}\right)$ and, hence, the thesis.
\end{proof}

\section{Layer-wise Sampling and Surrogate Objective}\label{s:surrogate}

\begin{figure}
    \centering
    \includegraphics[scale=0.8]{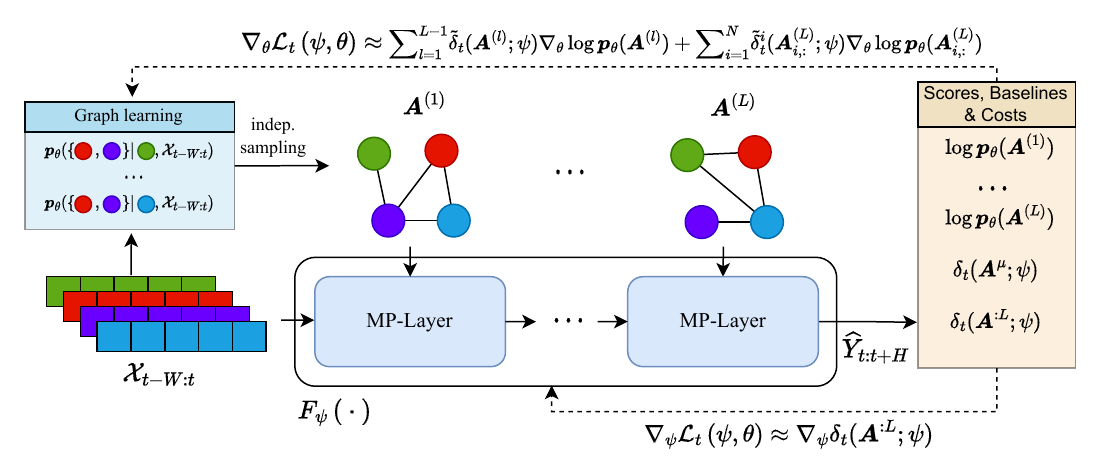}
    \caption{Overview of the learning architecture with layer-wise sampling and surrogate objective. The graph module samples a graph for each MP layer of predictor $F_\psi$.}
    \label{fig:framework-extended}
\end{figure}

As a final step, we can leverage on the structure of MP neural networks to rewrite the gradient $\nabla_\theta\losst$. This formulation allows for obtaining a different estimator for the case where we sample a different $\gso^{(l)}$ at each of the $L$ MP layers of $F_\psi$ (e.g., see Equation~\eqref{eq:stgnn-tts}). 
A schematic overview of the procedure is shown in Figure~\ref{fig:framework-extended} where $\gso^{:L}=\{\gso^{(l)}\}_{l=1}^L$.
\begin{proposition}\label{p:grad}
Consider family of models $F_\psi({}\cdot{}; \gso^{:L})$ with exactly $L$ message-passing layers propagating messages w.r.t.\ different adjacency matrices $\gso^{(l)}$, $l=1,\dots,L$, sampled from $\ptheta$~(either BES or SNS).
Assume that the cost function $\delta_t$ can be written as the summation over node-level costs $\delta^i_t$.
Then
\begin{align}\label{eq:loss-decomposition}
    \nabla_\theta \losst
    &= \E_{\ptheta} \left[\sum_{l=1}^{L-1}\delta_t(\gso^{:L}; \psi)\nabla_\theta \log\ptheta(\gso^{(l)}) + \sum_{i=1}^N\delta^i_t(\gso^{:L}; \psi) \nabla_\theta \log\ptheta(\gso^{(L)}_{i,:}) \right],
\end{align}
where $\gso^{(L)}_{i,:}$ denotes the $i$-th row of adjacency matrix $\gso^{(L)}$, i.e., the row corresponding to the neighborhood of the $i$-th node.
\end{proposition}
Proposition~\ref{p:grad} holds for all parametrizations of $\ptheta$ as long as the neighborhood of each node~(i.e., the rows of $\gso$) are sampled independently. 
Furthermore, note that almost all of the cost functions typically used for node-level tasks satisfy the assumption, e.g.,
\begin{align*}
    \widehat\mY_{t:t+H} = F_\psi\left(\gX_{t-W:t}; \gso^{:L}\right),&& \delta_t(\gso^{:L};\psi) = \sum_{i=1}^N \Big\lVert \vy^i_{t:t+H} - \widehat \vy^i_{t:t+H}\Big\rVert_p^p =  \sum_{i=1}^N\delta_t^i(\gso^{:L};\psi).
\end{align*}
The following provides proof of Proposition~\ref{p:grad} and presents a surrogate objective function inspired by Equation~\eqref{eq:loss-decomposition}.

\medskip
\begin{proof}
A proof can be derived by noticing the independence of $\delta_t^i(\gso^{:L}; \psi)$ and $\ptheta(\gso^{(L)}_{j,:})$ for $i \neq j$, and by exploiting the fact that with both BES and SNS rows of each $\gso^{(l)}$ are sampled independently. For the sake of readability, we omit the dependency of $\delta_t$ and $\delta_t^i$ from $\gso^{:L}$ and $\psi$. The proof follows:
\begin{align}
    \nabla_\theta \losst
    &= \E_{\ptheta} \left[\delta_t\nabla_\theta \log\ptheta(\gso^{:L}) \right]\\
    &= \E_{\ptheta} \left[\sum_{l=1}^{L-1}\delta_t\nabla_\theta \log\ptheta(\gso^{(l)})\right] + \underbrace{\E_{\ptheta} \left[\delta_t\nabla_\theta \log\ptheta(\gso^{(L)})\right]}_{ (*)}.
\end{align}   
By considering the second term:
\begin{align}
    (*) &= \E_{\ptheta} \left[\delta_t\nabla_\theta \log\ptheta(\gso^{(L)})\right]\\
    &= \E_{\ptheta} \left[\sum_{i=1}^N\delta^i_t  \sum_{j=1}^N\nabla_\theta\log\ptheta(\gso^{(L)}_{j,:}) \right]\\
    &= \E_{\ptheta} \left[\sum_{i=1}^N\delta^i_t \nabla_\theta \log\ptheta(\gso^{(L)}_{i,:}) \right] + \underbrace{\E_{\ptheta} \left[\sum_{i=1}^N\delta^i_t \sum_{j\neq i} \nabla_\theta \log\ptheta(\gso^{(L)}_{j,:}) \right].}_{(**)}
\end{align}
The two factors in $(**)$ are independent since $\delta_t^i$ depends only on $\gso^{:L-1}$ and $\gso^L_{i,:}$, hence 
\begin{align}
    (**) &=\sum_{i=1}^N\E_{\ptheta} \left[\delta^i_t \right]\sum_{j\neq i}\underbrace{\E_{\ptheta}\left[\nabla_\theta \log\ptheta(\gso^{(L)}_{j,:})\right]}_{=0} = 0.
\end{align}
Putting everything together, we get Equation~\eqref{eq:loss-decomposition} and the proof is completed. 
\end{proof}

\subsection{Surrogate Objective}

Intuitively, the second term in Equation~\eqref{eq:loss-decomposition} can be interpreted as directly rewarding connections that lead to accurate final predictions w.r.t.\ the local cost $\delta^i$. Besides providing a more general MC estimator, Preposition~\ref{p:grad} motivates us in considering a similar surrogate approximate loss $\surrlosst$ for the case where we use a single $\gso$ for all layers, i.e.,  we consider
\begin{equation}\label{e:approx2}
    \nabla_\theta \surrlosst = \E_{\ptheta}\Big[ \lambda{\delta}_t(\gso; \psi)\nabla_\theta \log\ptheta(\gso) + \small{\sum}_{i=1}^N{\delta}^i_t(\gso; \psi)\nabla_\theta \log\ptheta(\gso_{i,:})\Big],
\end{equation}
as gradient to learn $\ptheta$.
Equation~\eqref{e:approx2} is developed from Equation~\eqref{eq:loss-decomposition} by considering a single sample $\gso\sim \ptheta$ and introducing the hyperparameter $\lambda$.  Note that, in this case, $\surrlosst$ is an approximation of the true objective with a reweighting of the contribution of each $\delta^i(\gso; \psi)$. Following this consideration, $\lambda$ can be interpreted as a trade-off between local and global cost. In practice, we set $\lambda = 1/N$, so that the two terms are roughly on the same scale. Empirically, we observed that using the modified objective consistently leads to faster convergence; see Section~\ref{sec:exp}.

\section{Experiments}\label{sec:exp}

To validate the effectiveness of the proposed framework, we carried out experiments in several settings on both synthetic and real-world datasets. In particular, a set of experiments focuses on the task of graph identification where the objective is that of retrieving graphs that better explain a set of observations given a (fixed) predictive model. The second collection of experiments shows instead how the proposed approach can be used as a graph-learning module in an end-to-end forecasting architecture. 

\subsection{Datasets}

\begin{table}[ht]
\centering
\begin{tabular}{ l | c c c}
\toprule
\multicolumn{1}{c|}{Dataset} & \# nodes & \# edges & \# steps\\
\midrule
GPVAR & 30 & 98 & 30000\\
PEMS-BAY & 325 & 2369 & 52128\\
METR-LA & 207 & 1515 & 34272\\
AQI~(Beijing) & 36 & 180 & 8760\\
AQI~(Tianjin) & 27 & 135 & 8760\\
\bottomrule
\end{tabular}
\caption{Additional information on the considered datasets.}
\label{t:datasets}
\end{table}

We consider one synthetic dataset and $3$, openly available, real-world benchmarks.

\begin{itemize}

\item 
\textbf{GPVAR} -- The GPVAR synthetic dataset consists of signals generated by recursively applying a polynomial Graph VAR filter~\citep{isufi2019forecasting} and adding Gaussian noise at each time step: this results in complex, yet known and controllable, spatiotemporal dynamics. In particular, analogously to~\citet{zambon2022az}, we consider the data generating process
\begin{equation}
    \mX_t = \text{tanh}\left(\sum_{l=0}^L\sum_{q=1}^Q \Theta_{l,q}\widetilde{\gso}^l\mX_{t-q}\right) + \eta_t
\end{equation}
where $\widetilde{\gso} = \mI + \gso$ (with $\mI$ being the identity matrix), $\Theta \in \sR^{(L+1) \times Q}$ denotes the model parameter and $\eta_t \sim \mathcal N(0, \mI)$ is a Gaussian noise vector. Model parameters, with $L=Q=2$, are set as described in~\citep{zambon2022az} and used to generate a trajectory of $T=30000$ steps. We use $70/10/20\%$ data split for training, validation, and testing, respectively.

\item \textbf{AQI} -- The Air Quality Index~(AQI) dataset consists of hourly readings from air quality monitoring stations scattered over different Chinese cities. AQI has been previously used as a benchmark for time series imputation methods~\citep{yi2016st, cini2022filling, marisca2022learning}. We use the same preprocessing and data splits of previous works~\citep{yi2016st}. The ground-truth graph is obtained by considering the pairwise distance of the sensors, following the procedure used in~\citep{cini2022filling}.

\item
\textbf{METR-LA} and \textbf{PEMS-BAY} -- The METR-LA and PEMS-BAY datasets from~\citep{jagadish2014big, li2018diffusion} are two popular benchmarks in the traffic forecasting literature. The datasets consist of traffic speed measurements taken at crossroads in Los Angeles and San Francisco, respectively. We use the same preprocessing and data splits of previous works~\citep{wu2019graph}. The underlying graphs are extracted from the geographic position of the sensors following the same steps of~\cite{wu2019graph}.
\end{itemize} 

Additional relevant information about the datasets is provided in Table~\ref{t:datasets}.

\subsection{Controlled Environment Experiments}\label{sec:gpvar}

To gather insights on the impact of each aspect of the methods introduced so far, we start by using the controlled environment provided by the GPVAR dataset. 

\subsubsection{Graph Identification and Time Series Forecasting}\label{sec:gpvar-exp}

\begin{figure}[t]
    \centering
    \includegraphics[width=0.9\textwidth]{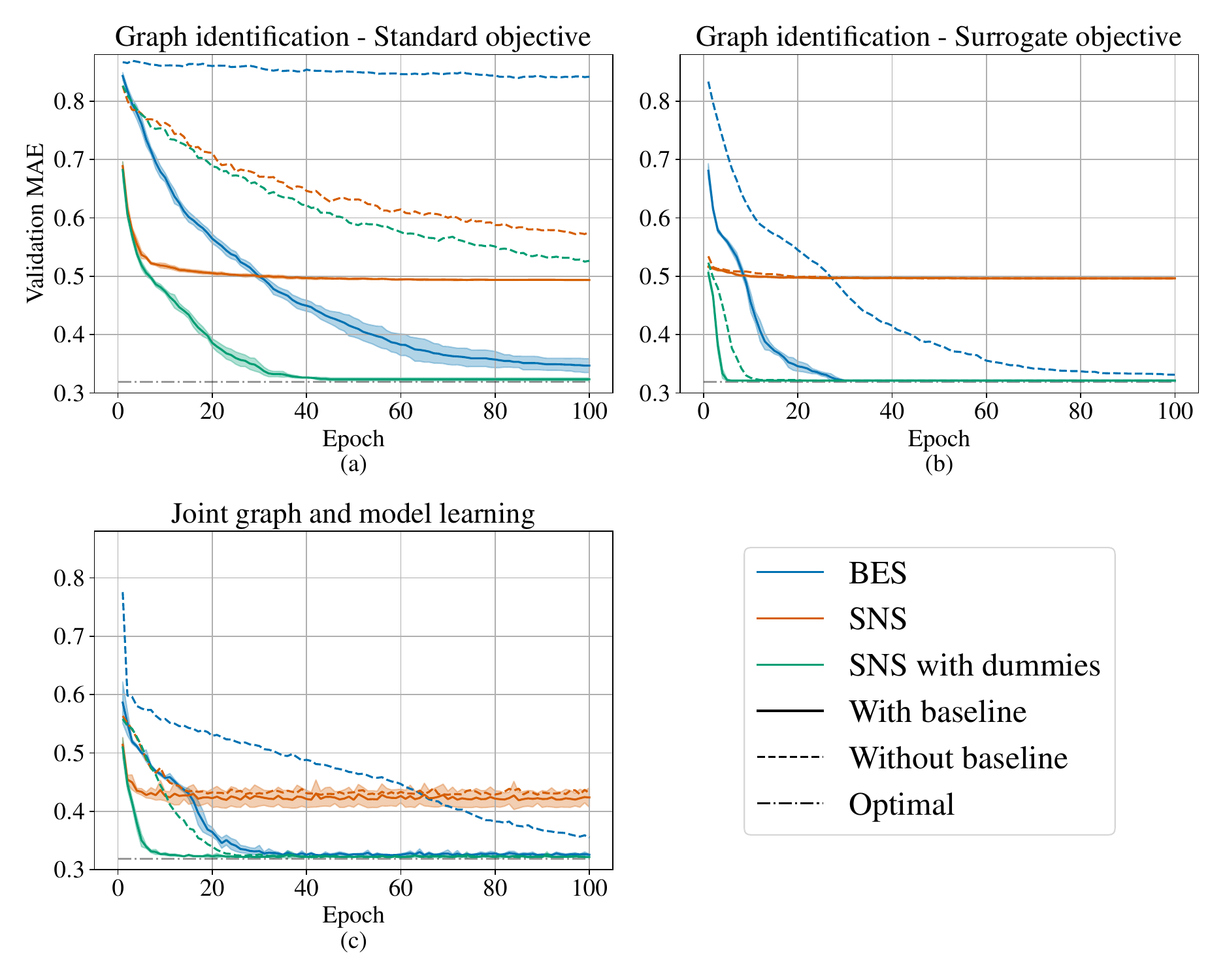}
    
    \begin{minipage}{0.32\textwidth}
    \phantomsubcaption{\label{fig:gpvar-a}}
    \end{minipage}
    \begin{minipage}{0.32\textwidth}
    \phantomsubcaption{\label{fig:gpvar-b}}
    \end{minipage}
    \begin{minipage}{0.32\textwidth}
    \phantomsubcaption{\label{fig:gpvar-c}}
    \end{minipage}
    \caption{Experiments on GPVAR. All the curves show the validation MAE after each training epoch. 
    }
    \label{fig:gpvar}
\end{figure}

In the first setup, we consider a GPVAR filter as the predictor and assume known the true model parameters, i.e., the coefficients of the filter, to decouple the assessment of the graph-learning module from that of the forecasting module.
Then, in a second scenario, we learn the graph while, at the same time, fitting the filter's parameters. Figure~\ref{fig:gpvar} shows the validation mean absolute error~(MAE) after each training epoch by using BES and SNS samplers, with and without baseline $\hat\beta$ for variance reduction, and when SNS is run with dummy nodes for adaptive node degrees. The number of maximum neighbors is set to $K=5$, which is the maximum degree of the ground truth graph.
In particular, Figure~\ref{fig:gpvar-a} and Figure~\ref{fig:gpvar-b} show results in the graph identification task for the vanilla gradient estimator derived from Equation~\eqref{eq:score-function-identity} and for the surrogate objective from Equation~\eqref{e:approx2}, respectively. To match the optimal prediction, models have to perfectly retrieve the underlying graph. During the evaluation, we used $\gsofr$ as input to the predictor instead of sampling $\ptheta$. Results allow us to make the following comments.

\begin{description}
\item[Impact of the Baseline]
The first striking outcome is the effect of baseline $\hat\beta$ in both the considered configurations which dramatically accelerates the learning process. 
\item[Graph distribution]
The second notable result is that, although both SNS and BES are able to retrieve the underlying graph, the sparsity prior in SNS yields faster convergence w.r.t.\ the number of samples seen during training, as the validation curves are steeper for SNS; note that the approximation error induced by having a fixed number of neighbors is effectively removed with the dummy nodes. 
\item[Surrogate objective]
Figure~\ref{fig:gpvar-b} shows that the surrogate objective contributes to accelerating learning even further for all considered methods. 
\item[Joint training]
Finally, Figure~\ref{fig:gpvar-c} reports the results for the joint training of the predictor and graph module with the surrogate objective. The curves, in this case, were obtained by initializing the parameters of the filter randomly and specifying an order of the filter higher than the real one; nonetheless, the learning procedure was able to quickly converge to the optimum when using as baseline the cost evaluated w.r.t.\ $\gsofr$.
\end{description}

\begin{figure}[t]
    \centering
    \includegraphics[width=0.8\textwidth]{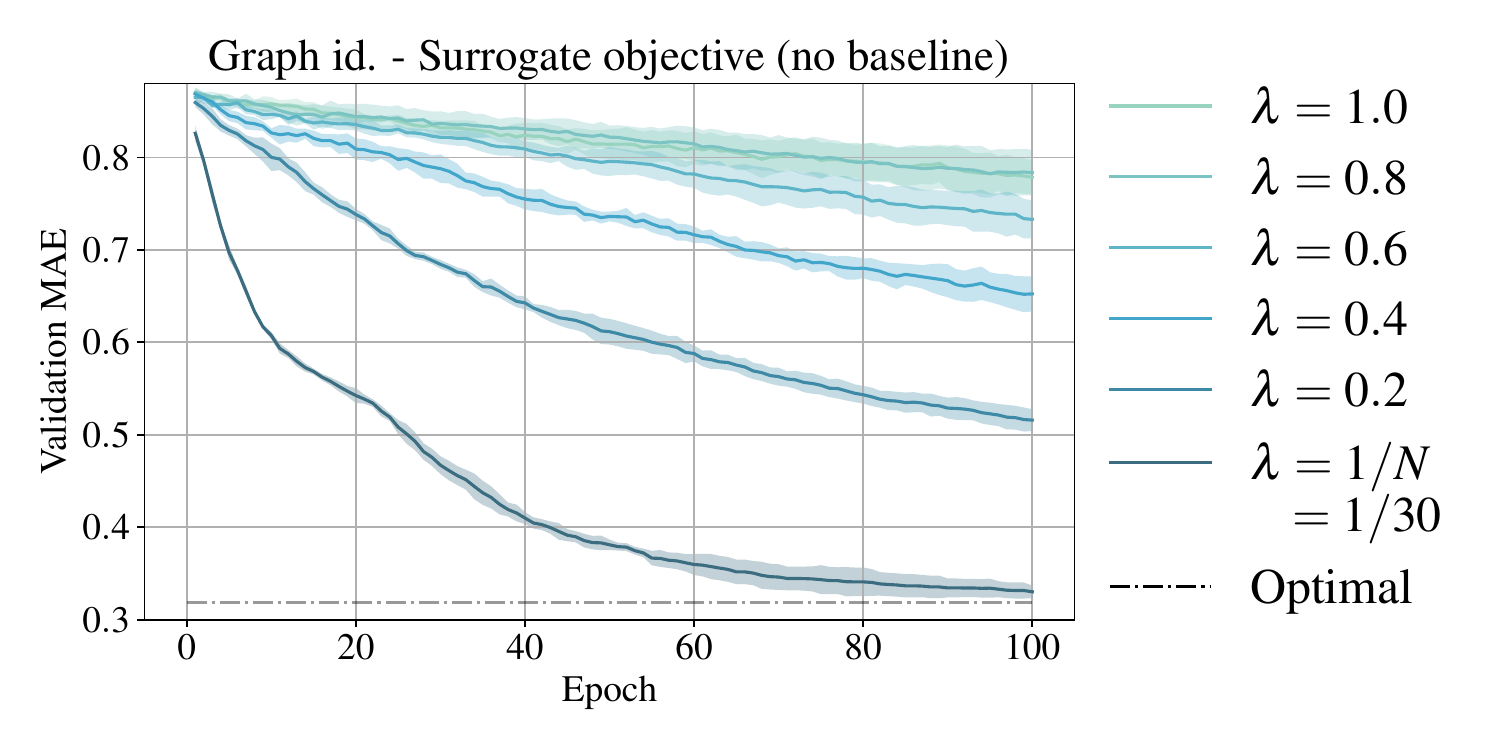}
    \caption{Sensitivity analysis on $\lambda$ for the surrogate objective.}
    \label{fig:lam}
\end{figure}

\subsubsection{Sensitivity Analysis}

To further assess the impact of the surrogate objective and that of the structural priors embedded into the SNS parametrization, we run a sensitivity analysis on both these aspects.

\begin{figure}[t]
    \centering
    \includegraphics[width=0.8\textwidth]{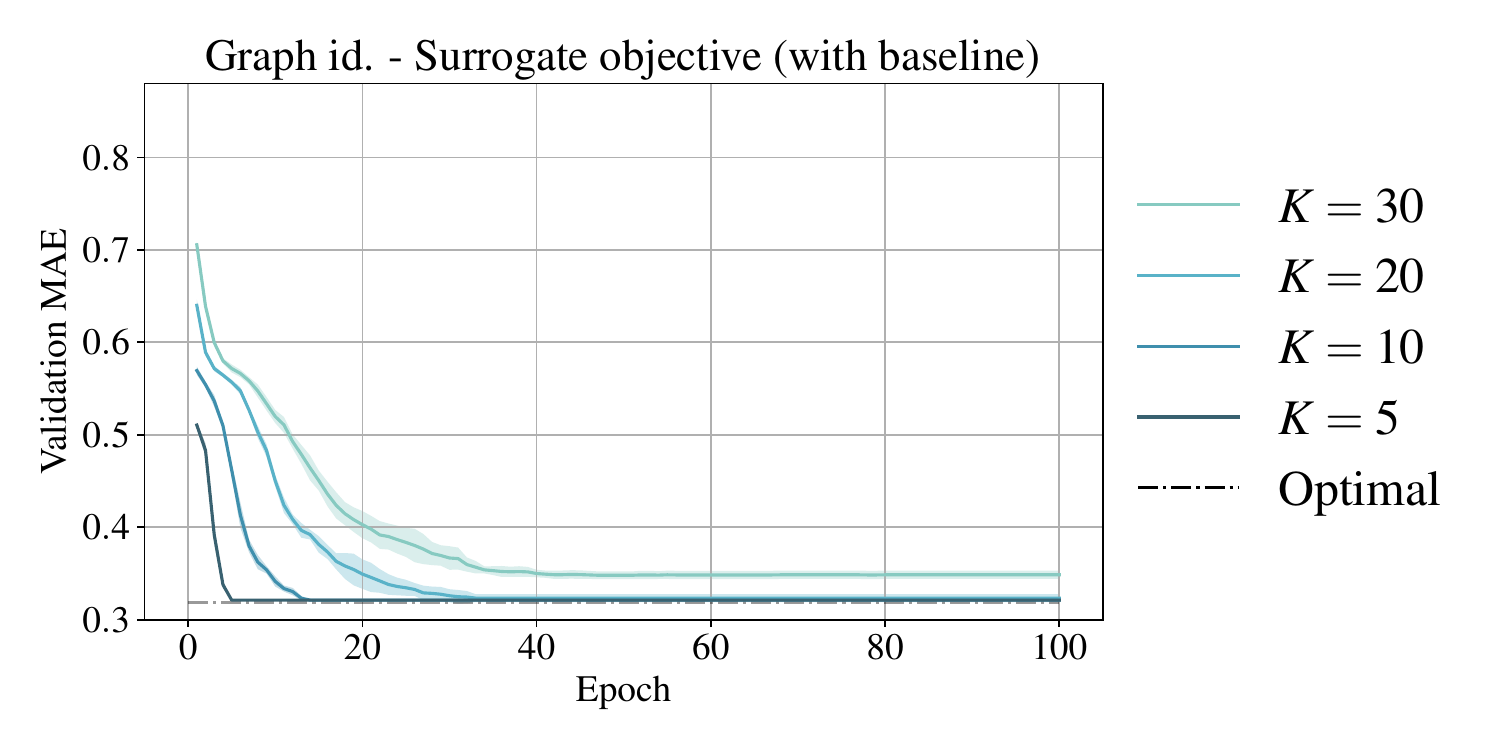}
    \caption{Sensitivity analysis on $K$ for SNS.}
    \label{fig:ks}
\end{figure}

Regarding the surrogate objective, we run a sensitivity analysis on the hyperparameter $\lambda$, which was kept fixed to $\lambda=1/N$ in the experiments in Figure~\ref{fig:gpvar}. In particular, we repeated the experiment on graph identification setting by considering the BES parametrization and values for $\lambda$ in the range $[1/N=1/30, 1]$. We did not use the baseline to accentuate the sensitivity to $\lambda$. Results, shown in Figure~\ref{fig:lam}, demonstrate the effectiveness of the surrogate loss in accelerating learning by introducing and reweighting the local cost term and how decreasing the weight of the global cost leads to faster convergence. 

Finally, we assess the impact of the value of the hyperparameter $K$ on the learning speed for the SNS sampler. In this case, we consider the graph identification experiment with the baseline for variance reduction. 
We run experiments with $K\in (5, 10, 20, 30)$ and a number of dummy nodes equal to $K-1$. Results in Figure~\ref{fig:ks} show that while the use of dummy nodes reduces the impact of a wrong assessment of $K$, overestimating the maximum number of neighbors can nonetheless lead to slower convergence. In particular, given these settings and hyperparameters, SNS fails to converge to the optimal solution for $K=30$, i.e., a number of neighbors equal to the number of nodes. As a general recommendation, we argue that using SNS can be beneficial as long as $K<N/2$, while for larger values of $K$ a BES parametrization is preferable due to the reduced overhead in sampling and likelihood evaluation.

\subsubsection{Comparison with Path-wise and Straight-through Estimators}

\begin{figure}[t]
    \centering
    \begin{minipage}{0.5\textwidth}
    \phantomsubcaption{\label{fig:gpvar-comp-est-a}}
    \end{minipage}
    \begin{minipage}{0.49\textwidth}
    \phantomsubcaption{\label{fig:gpvar-comp-est-b}}
    \end{minipage}
    \includegraphics[width=0.8\textwidth]{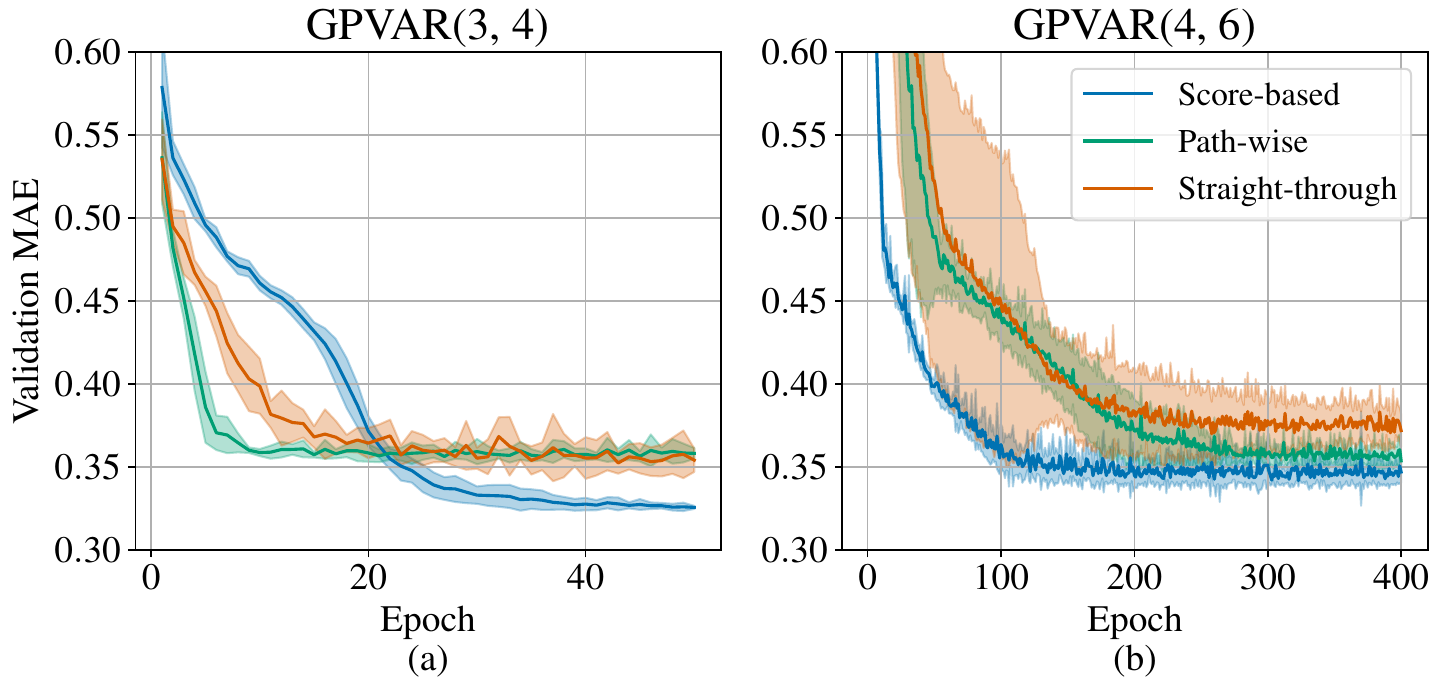}
    \caption{Comparison of different estimators on the joint training settings in GPVAR.}
    \label{fig:gpvar-comp-est}
\end{figure}

In this section, we assess the effectiveness of the proposed score-function estimator~(with baseline and surrogate objective) against both the path-wise estimator, based on the Concrete continuous relaxation of Bernoulli random variables~\citep{maddison2017concrete}, and the straight-through estimator~\citep{bengio2013estimating}. 
We consider the controlled joint graph and model learning scenario from Section~\ref{sec:gpvar-exp}.
In particular, for all estimators, we consider the BES parametrization for the graph distribution and the model family of Graph VAR filters of spatial order $3$ and temporal order $4$---as in the joint training experiment of Section~\ref{sec:gpvar-exp}---and a more difficult scenario corresponding to filters up to orders $4$ and $6$, respectively.

The results of the experiment are shown in Figure~\ref{fig:gpvar-comp-est}. In the simpler setting~(Figure~\ref{fig:gpvar-comp-est-a}), both the path-wise and straight-through estimators appear to converge faster than the score-based approach, yet they reach sub-optimal results---a side-effect that we attribute to the bias of the path-wise and straight-through estimators.
In the harder setting~(Figure~\ref{fig:gpvar-comp-est-b}), instead, our method achieves better performance both in terms of forecasting accuracy and sample complexity. This behavior might be associated with the complex dynamics of learning the relational structure given a larger family of predictive models. 

\subsection{Real-World Datasets}

The following discusses the application of the proposed method w.r.t.\ data coming from real-world scenarios.

\subsubsection{Graph Identification in AQI}
For graph identification, we set up the following scenario. From the AQI dataset, we extract $2$ subsets of sensors that correspond to monitoring stations in the cities of Beijing and Tianjin, respectively. We build a graph for both subsets of data by constructing a K-NN graph of the stations based on their distance; we refer to these as ground-truth graphs. Then, we train a different predictor for each of the two cities, based on the ground-truth graph. In particular, we use a TTS STGNN with a simple architecture consisting of a GRU~\citep{chung2014empirical} encoder followed by $2$ MP layers. As a reference value~(sanity check), we also report the performance achieved by a GRU trained on all sensors, without using any spatial information. Performance is measured in terms of 1-step-ahead MAE. 
\begin{wraptable}{r}{0.4\textwidth}
\centering
\resizebox{\linewidth}{!}{
\begin{tabular}{ l | r r }
\multicolumn{1}{c}{}&\multicolumn{2}{c}{Tested on}\\
\cmidrule[0.8pt]{2-3}
\multicolumn{1}{c}{Trained on}&\multicolumn{1}{|c}{Beijing}&\multicolumn{1}{c}{Tianjin}\\

\midrule
\text{Beijing} & 9.43 {\tiny $\pm$ 0.03} & 10.62 {\tiny $\pm$ 0.05}\\
\text{Tianjin}  & 9.55 {\tiny $\pm$ 0.06}  &10.56 {\tiny $\pm$ 0.03}\\
\midrule
\midrule
\text{Baseline}  & 10.21 {\tiny $\pm$ 0.01} & 11.25 {\tiny $\pm$ 0.04}\\
\midrule
\end{tabular}}
\caption{AQI experiment.}
\label{t:aqi}
\end{wraptable}%
Results for the two models, trained with early stopping on the validation set and tested on the hold-out test set for the same city~(i.e., in a transductive learning setting) are shown in the main diagonal of Table~\ref{t:aqi}. 
In the second stage of the experiment, we consider an inductive setting:  we train the model above on one of the two cities as a source, freeze its parameters, discard the ground-truth graph w.r.t.\ the left-out city, and train our graph learning module~(with the SNS parametrization) to maximize the forecasting accuracy. The idea is to show that our module is able to recover a topology that gives performance close to what would be achievable with the ground-truth graph.  Results, reported in the off-diagonal elements of Table~\ref{t:aqi}, show that our approach is able to almost match the performance that would have been possible to achieve by fitting the model directly on the target dataset with the ground-truth adjacency matrix; moreover, the performance is significantly better than that of the reference GRU.

\subsubsection{Joint Training and Forecasting in Traffic Datasets} 

Finally, we test our approach on $2$ widely used traffic forecasting benchmarks. Here we took the full-graph attention architecture proposed in~\citep{satorras2022multivariate}, removed the attention gating mechanism, and used the graph learned by our module to sparsify the learned attention coefficients; in particular, we considered the SNS sampler with $K=30$, $10$ dummy nodes and surrogate objective~($\lambda=1/N$). We used the same hyperparameters of~\citep{satorras2022multivariate}, except for the learning rate schedule and batch size~(see supplemental material). As a reference, we also tested results using the ground-truth graph, a graph with only self-loops~(i.e., with $\gso$ set to the identity matrix), as well as a random graph sampled from the Erd\"os-R\'enyi model with $p=0.1$. For MTGNN~\citep{wu2020connecting} and GTS we report results obtained by running the authors' code. More details are provided in Appendix~\ref{a:exp}. Note that GTS is considered the state of the art for methods based on path-wise estimators~\citep{zugner2021study}. Results in Table~\ref{t:traffic} show the MAE performance for $15$, $30$ and $60$ minutes time horizons achieved over multiple independent runs. 
Our approach is always competitive w.r.t.\ the state-of-the-art alternatives, and statistically better than all the baselines with reference adjacency matrices. Note that, using a random adjacency matrix---which essentially corresponds to randomly sparsifying the attention coefficients---is often competitive with more complex approaches which suggests that, in some datasets, having access to the ground-truth graph is not decisive for achieving high performance. That being said, our graph learning methods consistently improve performance w.r.t.\ the na\"ive baselines.

\begin{table}[t]
\centering
\resizebox{\linewidth}{!}{
\begin{tabular}{ l | r r r |  r r r}
\multicolumn{1}{c}{}&\multicolumn{3}{|c}{METR-LA} & \multicolumn{3}{|c}{PEMS-BAY}\\
\cmidrule[.8pt]{1-7}
 \multicolumn{1}{c|}{Model} & \multicolumn{1}{c}{\small MAE @ 15} & \multicolumn{1}{c}{\small MAE @ 30} & \multicolumn{1}{c|}{\small MAE @ 60} & \multicolumn{1}{c}{\small MAE @ 15} & \multicolumn{1}{c}{\small MAE @ 30} & \multicolumn{1}{c}{\small MAE @ 60}\\
\midrule
\text{Full attention} & 2.727 {\tiny $\pm$ .005} & 3.049 {\tiny $\pm$ .009} & 3.411 {\tiny $\pm$ .007} & 1.335 {\tiny $\pm$ .003} & 1.655 {\tiny $\pm$ .007} & 1.929 {\tiny $\pm$ .007}
\\
\midrule
\text{GTS} & 2.750 {\tiny $\pm$ .005} & 3.174 {\tiny $\pm$ .013} & 3.653 {\tiny $\pm$ .048} &1.360 {\tiny $\pm$ .011} & 1.715 {\tiny $\pm$ .032} & 2.054 {\tiny $\pm$ .061}\\
$\text{MTGNN}$  & 2.690 {\tiny $\pm$ .012} & 3.057 {\tiny $\pm$ .016} & 3.520 {\tiny $\pm$ .019} & 1.328 {\tiny $\pm$ .005} & 1.655 {\tiny $\pm$ .010} & 1.951 {\tiny $\pm$ .012}\\

\midrule
\text{Our (SNS)} & 2.725 {\tiny $\pm$ .005} & 3.051 {\tiny $\pm$ .009} & 3.412 {\tiny $\pm$ .013} & 1.317 {\tiny $\pm$ .002} & 1.620 {\tiny $\pm$ .003} & 1.873 {\tiny $\pm$ .005}
\\
\midrule
\midrule
{Adjacency}&\multicolumn{3}{|c}{}&\multicolumn{3}{|c}{}\\
\text{--Truth} & 2.720 {\tiny $\pm$ .004} & 3.106 {\tiny $\pm$ .008} & 3.556 {\tiny $\pm$ .011} & 1.335 {\tiny $\pm$ .001} & 1.676 {\tiny $\pm$ .004} & 1.993 {\tiny $\pm$ .008}\\
\text{--Random} & 2.801 {\tiny $\pm$ .006} & 3.160 {\tiny $\pm$ .008} & 3.517 {\tiny $\pm$ .009} & 1.327 {\tiny $\pm$ .001} & 1.636 {\tiny $\pm$ .002} & 1.897 {\tiny $\pm$ .003}\\
\text{--Identity} & 2.842 {\tiny $\pm$ .002} & 3.264 {\tiny $\pm$ .002} & 3.740 {\tiny $\pm$ .004} & 1.341 {\tiny $\pm$ .001} & 1.684 {\tiny $\pm$ .001} & 2.013 {\tiny $\pm$ .003}\\
\midrule

\end{tabular}}
\caption{Results on the traffic datasets.}
\label{t:traffic}
\end{table}

\subsection{Scalability}\label{sec:scalability}

\begin{figure}
    \centering
    \includegraphics[width=0.8\textwidth]{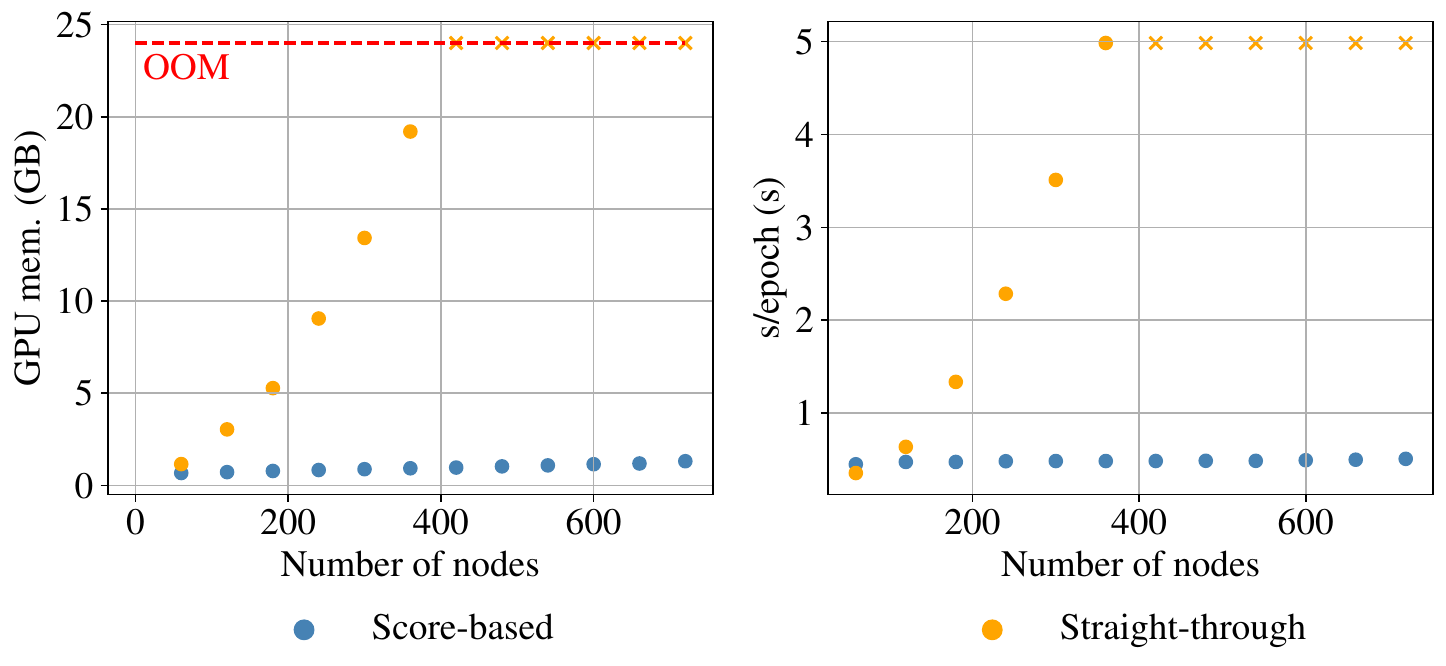}
    \caption{Computational scalability of the proposed estimator against the straight-through method.}
    \label{fig:scalability}
\end{figure}

To assess the scalability of the proposed method, we consider a T\&S model consisting of a message-passing GRU~(MPGRU, \citealt{cini2022filling}), i.e., a GRU with gates implemented by MPNNs. In particular, we consider a simple MP scheme s.t.
\begin{equation}
    \vz_t^{i,(l)} = \sum_{j \in \mathcal{N}(i)}\textsc{MLP}\left(\vz_t^{i,(l-1)}, \vz_t^{j,(l-1)}\right).
\end{equation}
The resulting model has a space and time complexity that scales as $\bigO(LTE)$. By considering the same controlled environment of the experiments in Section~\ref{sec:gpvar} and varying the number of nodes in the graph underlying the generated data, we empirically assessed the time and memory cost of learning a graph distribution with our SNS approach against the straight-through approach. Note that, while the straight-through estimator allows for a sparse forward pass at inference, the processing is nonetheless dense at training time---thus requiring $\bigO(LTN^2)$ time and space, instead of $\bigO(LTE)$.

The resulting models are trained on mini-batches of $4$ samples with a window size of $8$ steps for $50$ epochs, each consisting of $5$ mini-batches. The empirical results in Figure~\ref{sec:scalability} show measured GPU usage and latency for the above settings. The computational advantages of the sparse message-passing operations of our method are evident.

\section{Conclusions}\label{s:conclusion}
 
In this paper, we propose a methodological framework for learning graph distributions from spatiotemporal data. Our novel probabilistic framework relies upon score-function gradient estimators that allow us for keeping the computation sparse throughout both the training and inference phases. We then develop variance-reduction techniques for our method to obtain accurate estimates for the training gradient. 
The proposed graph learning modules are applied to the time series forecasting task where they can be used for both graph identification and as components of an end-to-end architecture.  
Empirical results support our claims, showing the effectiveness of the framework. Notably, we achieve forecasting performance on par with state-of-the-art alternatives, while maintaining the benefits of graph-based processing. 
Possible directions for future research include the assessment of the proposed method w.r.t.\ inference of dynamic adjacency matrices, distribution agnostic variance reduction methods, and, in particular, the design of advanced forecasting architectures to achieve accurate predictions at scale.  Furthermore, it would interesting to assess the combination of the proposed estimators with orthogonal variance reduction techniques~(e.g., \citealt{kool2020estimating}) and data-driven baselines. Finally, future works might investigate the application of the recently proposed implicit maximum likelihood estimators~\citep{niepert2021implicit, minervini2023adaptive} to the settings explored in this paper.

\section*{Acknowledgements}

This work was supported by the Swiss National Science Foundation project FNS 204061: \emph{HigherOrder Relations and Dynamics in Graph Neural Networks}. The authors wish to thank the Institute of Computational Science at USI for granting access to computational resources.

\appendix

\section*{Appendix}

\section{Deferred Proofs}\label{a:proofs}

This Appendix provides the proofs for Lemma~\ref{lemma:frechet-fun} and Lemma~\ref{lemma:pi-geq-pj}.

\subsection{Proof of Lemma~\ref{lemma:frechet-fun}}\label{a:proof-lemma:frechet-fun}
Note that for all $\gso,\gso'\in \gA \triangleq \{0,1\}^{N\times N}$  the Fr\'echet function $\frfun_H$ can be expressed as
\begin{equation}
\frfun_H(\gso') = \frfun_F(\gso') \triangleq \E_{\gso\sim\ptheta}\left[\lVert \gso - \gso'\rVert_F^2\right]
\end{equation}
w.r.t.\ the Frobenius norm, therefore we have also 
\begin{equation}
\min_{\gso'\in \gA} \frfun_H(\gso') = 
\min_{\gso'\in \gA} \frfun_F(\gso').
\end{equation}
Note now that 
\begin{align}
\frfun_F(\gso') &=  \E_\ptheta\left[\lVert \gso - \gso'\rVert^2_F\right] 
    =   \E_\ptheta\left[\lVert \gso \pm \gsomean - \gso'\rVert^2_F\right]
\\& =   \E_\ptheta\left[\lVert \gso - \gsomean\rVert^2_F\right] 
        + 2 \E_\ptheta\left[\langle \gso-\gsomean ,\gsomean - \gso'\rangle_F\right] 
        +   \E_\ptheta\left[\lVert \gsomean - \gso'\rVert^2_F\right]
\\& =   \E_\ptheta\left[\lVert \gso - \gsomean\rVert^2_F\right] 
        + 2 \underbrace{\langle \E_\ptheta[\gso]-\gsomean ,\gsomean - \gso'\rangle_F}_{=0} 
        +   \lVert \gsomean - \gso'\rVert^2_F.
\end{align}
Moreover, as the first term does not depend on $\gso'$, the minimum of $\frfun_F(\gso')$ is achieved at the minimum of 
\begin{equation}
\lVert \gsomean - \gso'\rVert^2_F=\sum_{i,j=1}^N(\gsomean_{i,j} - \gso_{i,j}')^2.
\end{equation}

\subsection{Proof of Lemma~\ref{lemma:pi-geq-pj}}
The neighborhood of each node $n$ is sampled independently from the others, so we derive the proof for a reference node $n$ and denote $\phi=\Phi_{n,:}$.

Note that, for every pair of node $i,j\in\sensorset$ and scalar $g\in\mathbb R$
\begin{align}
    \mathbb P(G_{\phi_i}\ge g) &\ge \mathbb P(G_{\phi_j}\ge g)\\
        \iff&e^{-e^{-(g-\phi_i)}}=\cdff_{\phi_i}(g) \le \cdff_{\phi_j}(g)=e^{-e^{-(g-\phi_j)}} \\
        \iff&\left({e^{-e^{-g}}}\right)^{e^{\phi_i}} \le \left({e^{-e^{-g}}}\right)^{e^{\phi_i}}.
\end{align}
Being ${e^{-e^{-g}}}<1$ and the $e^x$ monotone we obtain
\begin{align}
    \mathbb P(G_{\phi_i}\ge g) &\ge \mathbb P(G_{\phi_j}\ge g)
    \iff e^{\phi_i} \ge e^{\phi_j}
    \iff \phi_i \ge \phi_j. \label{eq:phii-ge-phij}
\end{align}
$P(\gso_{n,i}=1)$ can then be rewritten as
\begin{align}
    \mathbb P(\gso_{n,i}=1) 
    &=\mathbb P(G_{\phi_i}\in \topk\{G_{\phi_l}:l\in\sensorset\}) 
    =\mathbb P(G_{\phi_i}\ge \overline G)
    \\&=\int \mathbb P(G_{\phi_i}\ge g) \pdff_{\overline G}(g)\, dg  \label{eq:int-mu-ni}
\end{align}
with $\overline G$ being the random variable associated with the $K$-th largest realization in $\{G_{\phi_l}:l\in\sensorset\}$ and $\pdff_{\overline G}$ its p.d.f., we obtain
\begin{align}
    \mathbb P(\gso_{n,i}=1) &\ge \mathbb P(\gso_{n,j}=1) 
    \ \stackrel{(\text{Eq.~\ref{eq:int-mu-ni})}}{\iff}\  \mathbb P(G_{\phi_i}\ge g) \ge \mathbb P(G_{\phi_j}\ge g)
    \ \stackrel{\text{(Eq.~\ref{eq:phii-ge-phij})}}{\iff}\  \phi_i \ge \phi_j,
\end{align}
concluding the proof.

\section{Details on the Computation of the SNS Likelihood}\label{a:trapezoid}

In this appendix, we provide all the steps to obtain the rewriting of the likelihood on an SNS sample introduced in Equation~\eqref{eq:sns-score-rewriting}. The derivations provided here follow from the results of~\citet{kool2020estimating}. 
\begin{multline*}
    \begin{aligned}
    \ptheta(S_K|i) &=\mathbb{P}\left(\min_{i\in S_K} G_{\phi_i} > \max_{i\in {\sensorset\setminus S_k}} G_{\phi_i}\right)\\
    &=\mathbb{P}\left(\min_{i\in S_K} G_{\phi_i} > G_{\phi_{\sensorset\setminus S_k}}\right)\\
    &= \mathbb{P}\left(G_{\phi_i} > G_{\phi_{\sensorset \setminus S_k}}, \forall i \in S_K\right)\\
    &= \int_{-\infty}^{\infty} \pdff_{\phi_{\sensorset \setminus S_k}}(g) \mathbb{P}\left(G_{\phi_i} > g, \forall i \in S_K\right)\,d g\\
    &= \int_{-\infty}^{\infty} \prod_{i\in S_K}\left(1 - \cdff_{\phi_i}\left(g\right)\right)\pdff_{\phi_{\sensorset \setminus S_k}}(g) \,d g \\
    &= \int_{0}^{1} \prod_{i\in S_K} \left(1 - \cdff_{\phi_i}\left(\cdff^{-1}_{\phi_{\sensorset\setminus S_k}}(v)\right)\right)\,d v \quad {\scriptstyle \Big\{v = \cdff_{\phi_{\sensorset \setminus S_k}}(g)\Big\}}\\
    &= \int_0^1\prod_{i\in S_k}\Big(1 - v^{\exp(\phi_i-\phi_{\sensorset\setminus S_K})}\Big)\,d v\\
    &= \exp\left( b\right)\int_0^1 u^{\exp\left(b\right) - 1}\prod_{i\in S_k}\left(1 - u^{\exp(\phi_i  - \phi_{\sensorset \setminus S_k} +  b)}\right)\,du \quad {\scriptstyle \Big\{u = v^{\exp{\left(-b\right)}}\Big\}}\\
    &= \exp\left( \phi_{\sensorset \setminus S_K} + c\right)\int_0^1 u^{\exp\left(\phi_{\sensorset \setminus S_K} + c\right) - 1}\prod_{i\in S_k}\left(1 - u^{\exp(\phi_i + c)}\right)\,du \quad {\scriptstyle \Big\{c = b - \phi_{\sensorset \setminus S_K}\Big\}},
    \end{aligned}
\end{multline*}
which corresponds to the desired rewriting.

\section{Experiments Details}\label{a:exp}

All the code for the experiments has been developed in Python using the following open-source libraries:
\begin{itemize}
    \item PyTorch~\citep{paske2019pytorch};
    \item PyTorch Geometric~\citep{fey2019fast};
    \item Torch Spatiotemporal~\citep{Cini_Torch_Spatiotemporal_2022};
    \item PyTorch Lightning~\citep{Falcon_PyTorch_Lightning_2019};
    \item numpy~\citep{harris2020array};
\end{itemize}
furthermore, we relied on Neptune\footnote{\url{https://neptune.ai/}}~\citep{neptune2021neptune} for logging experiments. For GTS, we used the code provided by the authors\footnote{\url{https://github.com/chaoshangcs/GTS}} to obtain the results shown in the table, however we fixed a bug in the performance evaluation present in the official implementation\footnote{\url{https://github.com/chaoshangcs/GTS/issues/19}}.

Experiments were run on a cluster equipped with Nvidia Titan V and GTX 1080 GPUs. The code to reproduce the experiments of the paper is available online\footnote{\url{https://github.com/andreacini/sparse-graph-learning}}. 

\subsection{Synthetic Experiments}

For the graph identification experiments, we simply trained the different graph identification modules using the Adam optimizer with a learning rate of $0.05$ to minimize the absolute error. For the joint graph identification and forecasting experiment, we train on the generated dataset a GPVAR filter with $L=3$ and $Q=4$ with parameters randomly initialized and fitted with Adam using the same learning rate for the parameters of both graph filter and graph generator. To avoid numeric instability, scores $\Phi$ were soft-clipped to the interval $(-5, 5)$ by using the $tanh$ function.

\subsection{AQI Experiment} 

For the experiments on AQI we use a simple TTS model with a GRU encoder with $2$ hidden layers, followed by a GNN decoder with $2$ graph convolutional layers updating representations as:
\begin{equation}
    \mZ^{(l)} = \sigma\left(\mD^{-1} \gso \mZ^{(l-1)}\mW + \mV \mZ^{(l-1)}\right)
\end{equation}
where $\mW,\mV \in \sR^{d_z \times d_z}$ are learnable weight matrices and $\sigma$ is a nonlinear activation function (in particular we use Swish~\citep{ramachandran2017searching}). All layers have a hidden size of $64$ units. We use an input window size of $24$ steps and train for $100$ epochs the models with the Adam optimizer with an initial learning rate of $0.005$ and a multi-step learning rate scheduler. For the GRU baseline, we use a single recurrent layer of size $64$ followed by an MLP decoder with $1$ hidden layer with $32$ units. For the graph module, we use SNS with $K=5$ and $4$ dummy nodes and train with Adam with a learning rate of $0.01$ for $200$ epochs. At test time, we used models with weights corresponding to the lowest validation error across epochs.

\subsection{Traffic Experiment} 

As reported in the paper, we use the same architecture and hyperparameters of the full graph model of~\citet{satorras2022multivariate}, except for the gating mechanism which was removed for the graph-based baselines. We train the models for a maximum of $200$ epochs with Adam and an initial learning rate of $0.003$ and a multi-step scheduler (analogously to~\citet{satorras2022multivariate}. Note that we used an initial learning rate lower than the one used in~\citep{satorras2022multivariate} as we observed it was on average leading to better performance. 
In each epoch, we used $200$ mini-batches of size $64$ for all the model variations, except for the full-attention model for which --on PEMS-BAY-- we had to limit the batch size to $16$ due to GPU memory limitations. For the graph learning module, we used SNS with $K=30$ and $10$ dummy nodes. We also used a temperature $\tau=0.5$ to make the sampler more deterministic. During evaluation, we used the $\gsofr$ to obtain test-time predictions.

\bibliography{main.bib}

\end{document}